\documentclass[conference]{IEEEtran}
\usepackage{afterpage}

\ifCLASSINFOpdf
\else
\fi

\usepackage{graphicx}
\usepackage[cmex10]{amsmath}
\usepackage{amsfonts}

\usepackage{algorithm}
\usepackage{algorithmic}
\usepackage{placeins}

\usepackage{amsthm}

\usepackage{marginnote}
\usepackage{color}

\definecolor{gray}{gray}{0.4}
\usepackage[font=bf]{caption}
  \DeclareCaptionType{copyrightbox}
\usepackage{subcaption}

\newcommand{\tr}{^{\top}} 

\usepackage{hyperref}
\usepackage{footnote}

\DeclareMathOperator*{\argmin}{argmin}
\DeclareMathOperator*{\argmax}{argmax}

\newtheorem{defn}{Definition}

\renewcommand{\algorithmicrequire}{\textbf{Input:}}
\renewcommand{\algorithmicensure}{\textbf{Output:}}
\newcommand{\algorithmicinit}{\textbf{Initialize:}}

\DeclareCaptionLabelFormat{continued}{Figure #2 (Continued)}
\begin{document}
%
\title{Learning Representations for Outlier Detection\\ on a Budget}


\author{\IEEEauthorblockN{Barbora Micenkov\'a}
\IEEEauthorblockA{Aarhus University\\
Aarhus, Denmark\\
Email: barbora@cs.au.dk}
\and
\IEEEauthorblockN{Brian McWilliams}
\IEEEauthorblockA{Department of Informatics\\
ETH Z\"urich, Switzerland\\
Email: brian.mcwilliams@inf.ethz.ch}
\and
\IEEEauthorblockN{Ira Assent}
\IEEEauthorblockA{Aarhus University\\
Aarhus, Denmark\\
Email: ira@cs.au.dk}}


%


\maketitle

\begin{abstract}

The problem of detecting a small number of outliers in a large dataset is an important task in many fields from fraud detection to high-energy physics. Two approaches have emerged to tackle this problem: unsupervised and supervised. Supervised approaches require a sufficient amount of labeled data and are challenged by novel types of outliers and inherent class imbalance, whereas unsupervised methods do not take advantage of available labeled training examples and often exhibit poorer predictive performance.  We propose BORE (a Bagged Outlier Representation Ensemble) which uses unsupervised outlier scoring functions (OSFs) as features in a supervised learning framework. BORE is able to adapt to arbitrary OSF feature representations, to the imbalance in labeled data as well as to prediction-time constraints on computational cost.
We demonstrate the good performance of BORE compared to a variety of competing methods in the non-budgeted and the budgeted outlier detection problem on 12 real-world datasets.
\end{abstract}



%
\IEEEpeerreviewmaketitle

\section{Introduction}
The aim of outlier or anomaly detection is to identify points in a dataset which deviate in some way from the usually observed patterns. Outliers typically represent a small portion of the data and can be very diverse in nature.

In applications where the semantics of outliers are known in advance (e.g., detection of fraud, intrusions or mislabeled data), labeled training examples for \emph{supervised} outlier detection may be available \cite{ChaBanKum09}.
Because outliers are naturally scarce, such data will be heavily imbalanced which poses a problem for most classifiers. 
Another challenge for supervised outlier detection is the heterogeneity of the outlier class. This makes generalising from a small number of labeled samples difficult. Furthermore, supervised methods are unable to detect novel types of anomalies for which no labeled training examples have been collected. 

\emph{Unsupervised} algorithms for outlier detection (see e.g.\ \cite{Agg13} for a review) are suitable for purely exploratory tasks where very few or no labeled examples of outliers are available.
These methods are based mainly on geometric properties of the data and typically assign a real-valued ``outlierness'' score to each data point. As such unsupervised algorithms are suited to detecting new types of outliers. However, they often exhibit poor predictive performance \cite{jair2013-semisup}. 

Recently, semi-supervised methods which modify unsupervised approaches to take advantage of labeled examples have been shown their promise \cite{jair2013-semisup}. 

In this work we take an entirely different approach to combining the strengths of unsupervised and supervised outlier detection: we propose a supervised algorithm that first learns a \emph{feature representation} which successfully differentiates outliers from inliers using the unlabeled data. Our method is not only an entirely novel approach that is able to make use of both unsupervised and supervised information, but also a simple and easily generalizable framework that allows incorporation of different outlier detection methods. It avoids the considerable effort in tuning parameters in existing work (e.g. kernels for non-linear feature transformations), and handles class imbalance in a straightforward fashion. Its final output can be easily interpreted as outlier probabilities. Finally, our algorithm can adapt to computational budgets at prediction time, providing good detection performance within user defined budget constraints.

Our contribution is two-fold: 
\begin{enumerate}
	\item  We propose a Bagged Outlier Representation Ensemble (BORE); a unified framework for incorporating unsupervised and supervised data for outlier detection. BORE first learns a representation of the outlierness of each point in an unsupervised fashion which are then used as features in a classifier trained on imbalanced data. 
	\item We consider the case where computational resources at prediction time are limited and introduce a feature selection technique that respects a computational budget while retaining good predictive performance. 
\end{enumerate}

The key idea underlying BORE is to view the output scores of unsupervised outlier scoring function (OSF) algorithms as non-linear transformations of the original feature space. Crucially, this new set of features provides a richer representation which better distinguishes outliers from normal points. This representation is then used for supervised learning, where we adopt an ensemble approach that elegantly handles class imbalance. Thus, an advantage of BORE is that its performance can be boosted by including a larger or more diverse set of outlier detectors -- particularly those which are known to be suited to the task at hand. This idea complements a recent line of research on \emph{outlier ensembles} that strives to combine outlier scoring functions \cite{Agg-posPaper,Zimek-posPaper} in an entirely unsupervised manner.

Since each feature is the output of an OSF learned on the whole dataset, adding features is costly. While this can be easily parallelized at training time, at prediction time, this issue is compounded with the fact that OSFs need to be run on the entire dataset including the new unseen test points in order to make predictions. Clearly this can be prohibitively expensive for large numbers of OSFs. This is particularly important when the prediction must be performed under time or computational constraints such as a real-time or embedded system.  To overcome this, we introduce a budget-aware feature selection approach which identifies a small subset of the OSFs that represent the best tradeoff between computational budget and prediction accuracy. Therefore at prediction time we obtain good performance which only requires computing a subset of features. This is a considerable improvement over existing approaches to representation learning that require large amounts of resources at both training and prediction time.



\textbf{Paper outline.} In the following section we briefly review current approaches to unsupervised and supervised outlier detection. We then detail BORE, our novel approach for learning representations for outlier detection. In order to deal with  computational budgets at prediction time we then propose a budget-aware variable selection procedure. 
Finally, we present extensive empirical results on 12 real-world datasets demonstrating the predictive outlier detection performance of BORE both with and without budget constraints.


\section{Related Work} \label{sec-rel}

\textbf{Feature representations.} 
In recent years, the field of \emph{representation learning} \cite{bengio2013representation} has become increasingly popular. In particular, a wide class of techniques based on deep neural networks have been proposed which learn rich feature representations of input data in a supervised or unsupervised fashion which can then be used for prediction. Such feature learning approaches have become the keystone of achieving state-of-the-art performance in a variety of problem domains. Elsewhere, less complex feature representations can be learned using correlations between features which have been shown to greatly improve prediction when few labeled examples are available \cite{mcwilliams2013correlated}. However, these methods require vast amounts of training data to learn good representations which are typically not available in outlier detection problems.

The key idea in feature learning is that good representations of the data can be obtained by means of solving an unsupervised learning problem. In this work, we leverage this idea with the addition of domain knowledge encoded in the zoo of existing specialised OSFs. 
In this respect, training OSFs on a particular dataset and using their outputs as input to a supervised learning problem can be viewed as unsupervised learning of a suitable representation for many possible types of outliers.

\textbf{Outlier Detection and Class-Imbalance Learning.} Outlier detection naturally faces the problem of class imbalance. Therefore, for the supervised case, well-established approaches from class-imbalance learning can be adopted including sampling, bagging and boosting, one-class classification and cost-sensitive learning (see e.g. in \cite{Agg13,combat-imbal}). Our proposed technique is not dependent on any of these models and can readily complement each of them. 


\textbf{Outlier Ensembles.} An appropriate combination of multiple unsupervised outlier scoring functions into ensembles can increase outlier detection performance \cite{Agg-posPaper,Zimek-posPaper}. However, building an ensemble is difficult in completely unsupervised settings and only heuristic approaches have been proposed so far \cite{SchWojZimKri12, nguyen_ensemble}. Open questions concern, e.g., the tradeoff between accuracy of single outlier scoring functions and their diversity or the normalization and the combination function of the outlier scores \cite{Zimek-posPaper}. 
No supervised approaches have been studied yet in this context except for an initial idea for a semi-supervised ensemble presented in \cite{learnOutEns}.
Our proposed approach could be viewed as an ensemble selection technique guided by the available training data, providing an elegant solution to the above stated problems. 

\textbf{Semi-supervised Outlier Detection.}
Semi-supervised techniques make use of both labeled and unlabeled data for training.
Recently, a semi-supervised anomaly detector (SSAD) was proposed\,\cite{jair2013-semisup}. 
While BORE incorporates the unsupervised information in its features, SSAD is based on an unsupervised technique, support vector data description \cite{svdd}, which learns a hypersphere enclosing the normal data, and uses this as a regularizer for supervised learning. 
It requires to specify an appropriate kernel on input.
The goal is, similarly to BORE, to achieve a good performance using labels while retaining the possibility of revealing novel anomalies through the unsupervised information. We experimentally compare BORE to this technique.

\section{Learning a Representation\\ for Outlier Detection} \label{sec-met}

\begin{figure*}[!t]
\begin{center}
\begin{subfigure}[b]{0.32\textwidth}	
\includegraphics[width=1\textwidth]{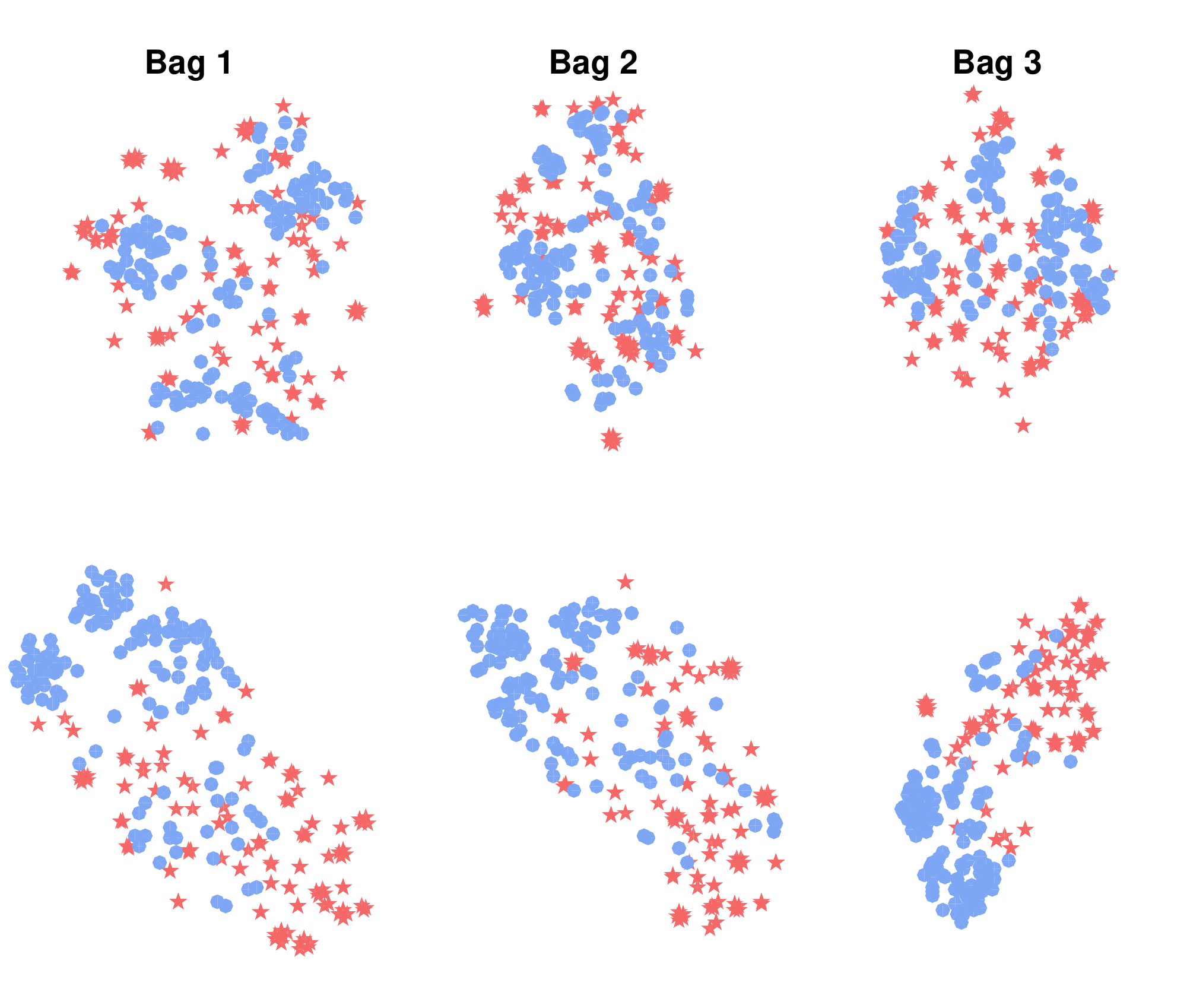}
\subcaption{Letter.} \label{sub:or_letter}
\end{subfigure} 
\begin{subfigure}[b]{0.32\textwidth}	
\includegraphics[width=1\textwidth]{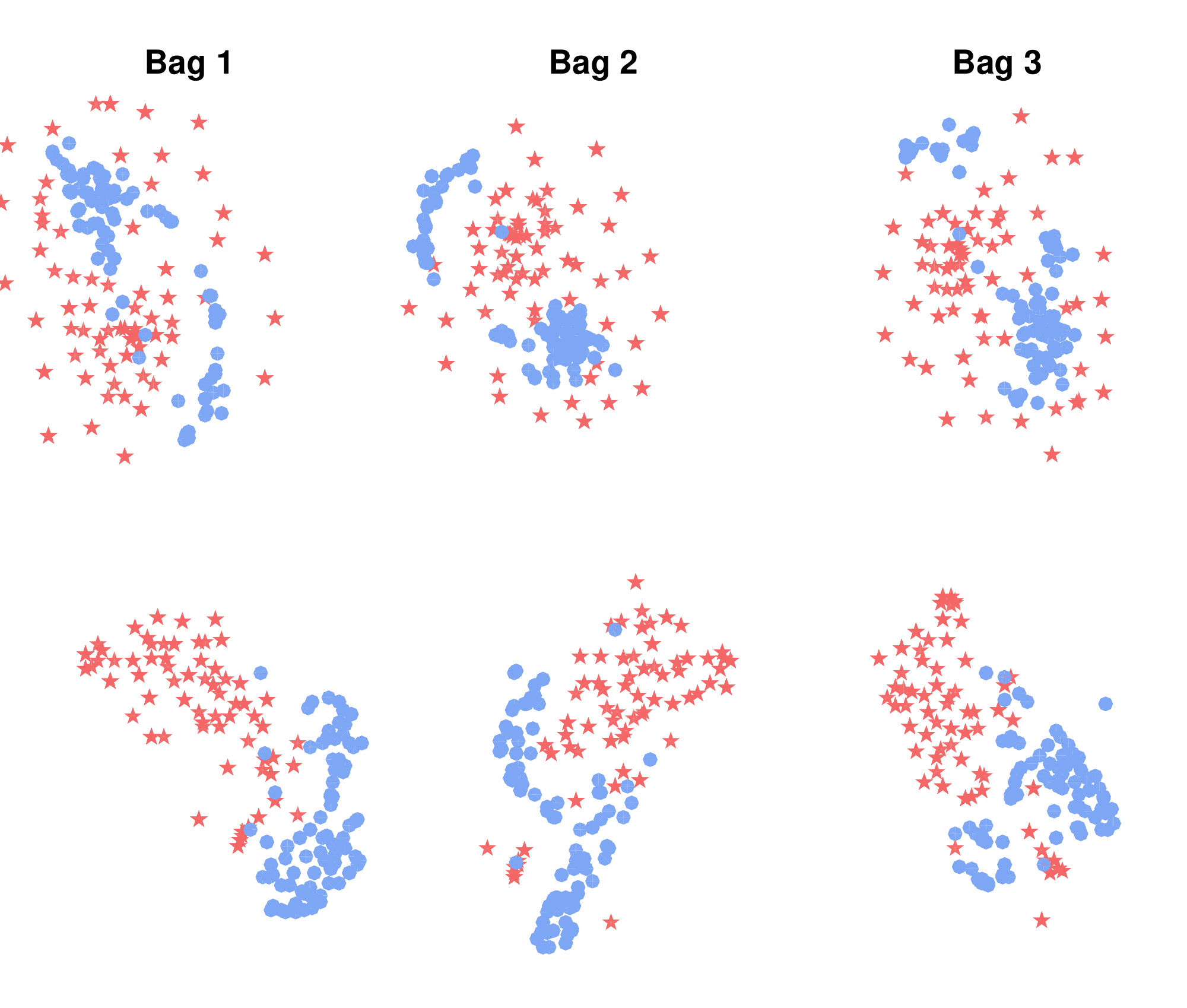}
\subcaption{Ionosphere.} \label{sub:or_iono}
\end{subfigure}
\begin{subfigure}[b]{0.32\textwidth}	
\includegraphics[width=1\textwidth]{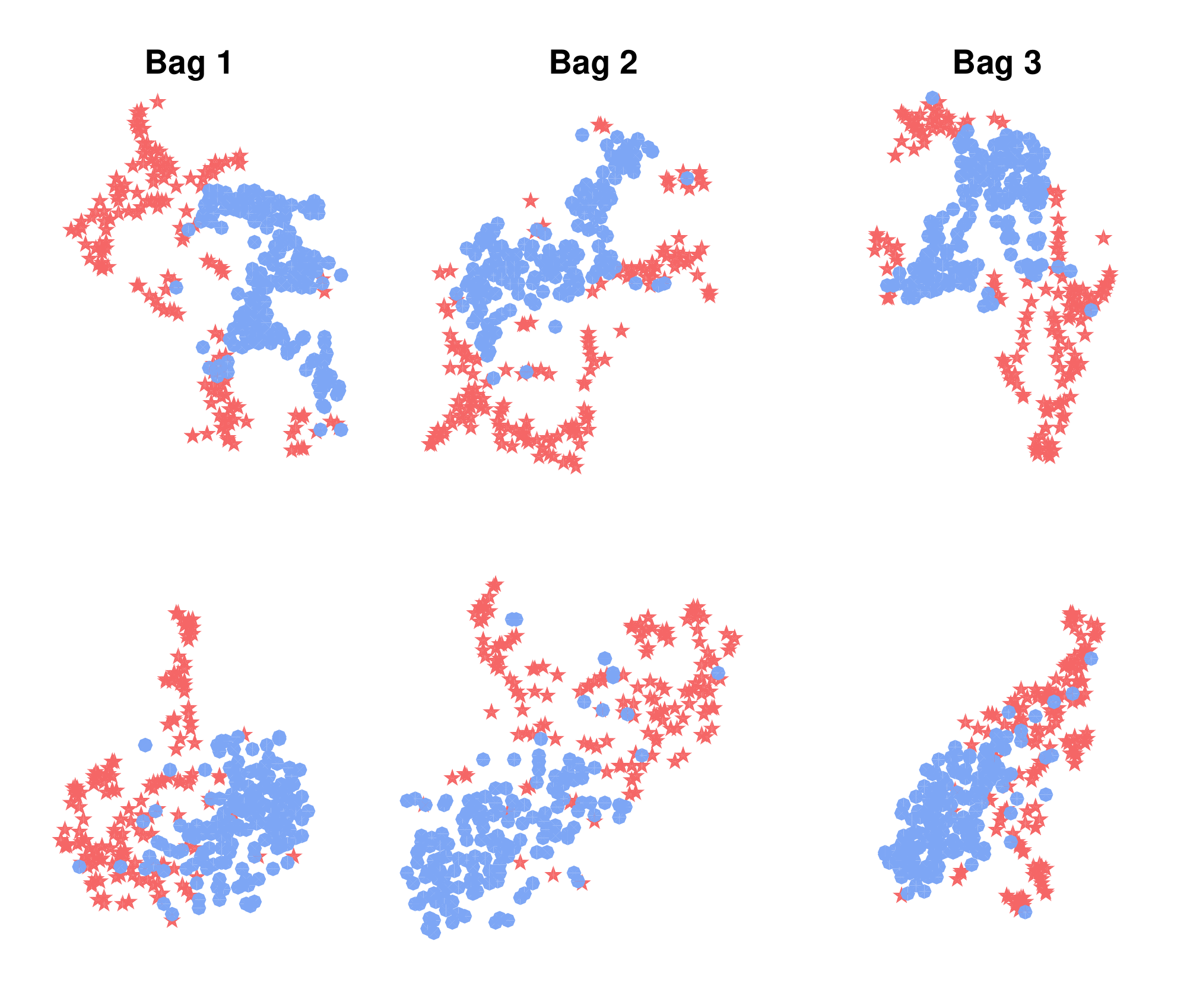}
\subcaption{Pageblocks.} \label{sub:or_page}
\end{subfigure} 
\caption{t-SNE visualisation of three different bags each subsampled from the (\subref{sub:or_letter}) Letter, (\subref{sub:or_iono}) Ionosphere and (\subref{sub:or_page}) Pageblocks data sets. The top line shows the original features and the bottom line shows the corresponding learned outlier representation. In each bag, the outlier representation exhibits a better separation between outlying points (shown in red) and non-outlying points (blue).}
\label{fig:tsne_letter}
\end{center}
\vspace{-0.5cm}
\end{figure*}




In this section we will detail our basic framework for outlier detection. First we describe the feature space construction which we use to learn a good representation of the outliers in the data in an unsupervised manner. We then use these features as input to a supervised learning procedure which adapts to the heterogeneity and class imbalance inherent in the outlier detection problem.

\subsection{Outlier Scoring Functions}
An outlier is broadly characterized as a point that deviates in some way from the rest of the data. The exact form of deviation depends on the data and the application.
Diverse detection and scoring functions have been proposed including approaches based on statistical methods, PCA, and other subspace analysis methods \cite{BarLew94, Jolliffe1986,MSS11,VuGA11}. A large body of work is based on the analysis of distances and density around data points, e.g. via $k$NN distance or relative local density \cite{RamRasShi00,BreKriNgSan00}. Our method supports any such detector that provide outlier scores.

For the discussion that follows, we require the following definition of an \emph{outlier scoring function} which will form the basis of our feature representation.

\begin{defn}[Outlier Scoring Functions]  \label{def:osf}
Given a data matrix $X\in\mathcal{X}\subseteq\mathbb{R}^{n\times k}$, an outlier scoring function (OSF) is a mapping $\Phi: \mathcal{X} \rightarrow  \mathbb{R}^n$. That is, an OSF assigns a real valued output to each row of the data matrix corresponding to the degree of outlierness of each point. 
\end{defn}
The scale and interpretation of different scoring approaches may vary. For example, they may be normalised between 0 and 1 so as to be interpreted as a probability or may be thresholded to assign binary labels to points. This makes standard ensemble approaches to combining OSFs difficult and highly dependent on scaling and normalization.

\subsection{Feature Space Construction}
Individually, the OSFs detailed above are limited in their ability to detect multiple types of outliers. The key insight we provide is that when combined, the output of multiple outlier detectors provide a good feature representation which can be used to detect many outlier types. 

Let $\Phi = \{\Phi_1,\dots,\Phi_m\}$ be a set of outlier scoring functions as in Definition \ref{def:osf}. Each $\Phi_j\in\Phi$ is applied to the data, $X$ and returns a vector $\Phi_j(X)\in\mathbb{R}^n$.
Our feature representation is then
\begin{equation}  
\Phi(X)=
\begin{bmatrix}
	\Phi_1(X) 
	,\cdots, 
	\Phi_m(X) \\
\end{bmatrix} . \label{eq:feattrans}
\end{equation}
\noindent Instead of the original data set, we now work with the transformed data set $\Phi(X)$ in the OSF feature space. Each individual $\Phi_j(X)$ can be viewed as a feature vector of the data.

To construct the set of functions $\Phi$, we may use any existing unsupervised OSF where the goal is to capture diverse aspects of the outliers in a particular dataset. It is therefore beneficial to expand the feature space by applying each OSF under a set of perturbations. For example multiple parameter settings, different distance metrics and different subspaces of the original features. 

In practise the original features of the data also contain useful information for identifying outliers. Therefore, we will use an augmented version of \eqref{eq:feattrans} where 
\begin{equation}\Phi(X)= [ X, \Phi_1(X),\ldots,\Phi_m(X) ] \in\mathbb{R}^{n\times d}\label{eq:OR}\end{equation} 
where the combined feature space dimension is $d=(k+m)$. With some abuse of notation we will denote $\Phi^{(i)}\in\mathbb{R}^{d}$ as the $i^{th}$ row of the matrix $\Phi(X)$ and $\Phi_j\in\mathbb{R}^n$ as the $j^{th}$ column. 

Using the original data and OSF features, we exploit the strength of unsupervised outlier detection to detect novel types of outliers within a supervised framework. To this end we will refer to the combined feature space in \eqref{eq:OR} as an \emph{outlier representation} (OR).

\subsection{Learning Setup}
We are now ready to incorporate supervised information on labeled outliers into our method.
The outlier representation learned in the previous section is a highly non-linear transformation of the original space. As such we can use a linear classifier in this new space to detect outliers. To make the discussion concrete, we adopt logistic regression, but in practise any linear classifier could be used instead.

Logistic regression (see e.g.\ \cite{hastie2009elements}) models the probability of a point $i$ being an outlier by means of a binary random variable, $y^{(i)}\in\{0,1\}$ conditioned on the outlier representation $\Phi^{(i)}$ and a parameter vector, $\beta\in\mathbb{R}^{d}$ through the logistic function:
\begin{equation}
  p(y^{(i)}=1|\Phi^{(i)};\beta) = \frac{1}{1+\exp{(-\beta^{\top} \Phi^{(i)})}} = \sigma(\beta^{\top} \Phi^{(i)}),
\label{eq:logreg}
\end{equation}
predicting 1 if $\sigma(\beta^{\top} \Phi^{(i)})>0.5$ and $0$ otherwise. The maximum likelihood estimator for $\beta$ is the solution to 
\begin{equation}
 \hat{\beta} = \argmin_{\beta} 
  - \sum_{i=1}^n  \log  \sigma(\beta^{\top} \Phi^{(i)})^{y_i} (1-\sigma(\beta^{\top} \Phi^{(i)}))^{(1-y_i)},
\label{eq:logregML}
\end{equation}
which can be solved efficiently using gradient descent or a pseudo second-order method such as L-BFGS \cite{hastie2009elements}.

In practise, any classifier can be used in place of logistic regression. However, since we already learn a highly non-linear feature representation, there is little additional utility to be gained by using a non-linear classifier compared with the additional cost in optimization and hyper-parameter tuning. This is illustrated in Section \ref{sec:results1} where we present results comparing BORE with a non-linear method, SSAD.

Another benefit of logistic regression in the context of outlier ensembles is that the output is easily interpreted since it directly models the probability of outlierness for a given datapoint given the outlier representation.



\noindent \newline {\bf Re-sampling and Bagging.} \label{sec:bag}
There are two challenges for the straightforward application of a standard classifier to outlier detection: The inherent class imbalance problem and the heterogeniety of the outlier class. We deal with both of these issues in a unified manner by adapting a standard re-sampling method: bootstrap aggregating, or \emph{bagging}  \cite{Breiman-bag}. Bagging constructs $B$ bags consisting of uniformly subsampling datapoints with replacement. It then averages the output of the models trained on each of the bags.  

Since uniform sampling would result in bags containing very few outliers, instead we use biased sampling to construct bags with balanced classes. For every bag $b=1,\ldots,B$, an equal number of outliers and inliers are sampled uniformly from their respective populations. Since the number of outliers is small compared with size of the inlier class, the same outliers will be appear in multiple bags while the inlier class will be substantially different between bags. 


Due to the inhomogeneity in the data, a different aggregation scheme such as Stacking \cite{breiman1996stacked} or  maximin aggregation  \cite{buhlmann2014magging} could be considered. However, we found empirically that neither performed as well as standard bagging.



\begin{table*}[!t]
\begin{center}
\small
\caption{Datasets used for evaluation. Outlier ratio in \% ($r$), dataset size ($n$), dimensionality ($d$).}
\label{tab:datasets}
\begin{tabular}{l r r r | l r r r | l r r r | l r r r }
\hline
Name & $r$ &     $n$ & $d$ & Name & $r$ & $n$ & $d$ & Name & $r$ & $n$ & $d$ & Name & $r$ & $n$ & $d$\\
\hline
{Cardio}      & 5&      $1734$ & $21$
&{Heart Disease}   & 10 &    $166$   & $13$
&{Hepatitis} & 9& 74& 19
&{Higgs} &5 &  $5000$ & $30$ \\

{Ionosphere}        &36 &    $351$  & $32$
&{Letter}  &6 &  $1600$   & $32$
&{Pageblocks}      &5 &  $5139$  & $10$
&{Parkinson} &20 &    $60$   & $22$ \\

{Pima}   &5 &    $526$  & $8$ 
&{Spambase}    &2 &   $2579$  & $57$
&{Waveform} & 3 &   $3443$   & $21$
&{Wilt} & 5 & $4839$ & $5$    \\
\hline 
\end{tabular}
\end{center}
\end{table*}

\begin{algorithm}[H] \caption{Orthogonal Matching Pursuit.} 
\label{alg:OMP}
	\begin{flushleft}
	\vspace{-0.25cm}
	\algorithmicrequire\;
	Data: $\{\Phi\in\mathbb{R}^{n\times d}, y\in\mathbb{R}^{n} \}$, \# non-zeros: $\gamma$ \\
	\algorithmicinit\; 
	\vspace{-0.25cm}
	$\mathcal{A} = \{\}, ~ R^0=y, ~ \beta^0 = {\mathbf 0}$
	\end{flushleft}
  \begin{algorithmic}[1] \label{alg:omp}
    \FOR {$t=1\ldots \gamma$}
           \STATE $j = \argmax_{j\notin \mathcal{A} } \frac{| \Phi_j \tr R^{t-1} | }{\Phi_j\tr \Phi_j}$
           \STATE $\mathcal{A} \leftarrow \{j\} \cup \mathcal{A}$
           \STATE $\beta^t_{\mathcal{A}}:$ solve \eqref{eq:logregML} with $\Phi_{\mathcal{A}}$. 
           \STATE $R^{t} = y -  \sigma({\beta^{t}_{\mathcal{A}}} \tr \Phi_{\mathcal{A}})$    
           \ENDFOR
  \end{algorithmic}
  \algorithmicensure\; $\beta^\gamma$
\end{algorithm}

\section{Outlier Detection on a Budget}

BORE learns a representation of arbitrary types of outliers by combining the output of many different outlier detectors in a bagging framework. Crucially, when the original dimensionality and the number of samples is large, learning this representation may be computationally intensive. This is typically not a problem since learning the outlier representation and training classifiers on individual bags can easily be computed in parallel. However, at test time when new instances must be classified this computational burden can be problematic. 

As a concrete example, the pre-trained outlier detection system might be deployed on less powerful hardware or might have to classify a point as an outlier under time constraints. The more OSFs used in the representation, the more resources required at test time to compute the representation of the new points.

To overcome this problem, we propose a classification strategy under a \emph{computational budget}.

\subsection{Cost-aware Feature Selection}
Associated with each feature transformation $\Phi_j(X)$ is a computational cost $c(j)$. For example, this cost could be directly related to the time or space complexity required to compute a particular feature.  
Some of the features may be too computationally expensive to generate at test time and equivalent predictive performance might be obtainable by instead combining a number of cheaply computed features. Our goal is good detection of outliers for any budget $C$ such that the sum of costs of utilised features $\sum_{j\in\mathcal{A}} c(j) \leq C$. 
We select features that are expected to have high utility for the outlier detection task while incurring low cost. In order to do so we adopt a strategy that also handles potential issues when the number of OSF features grows.

A challenge is statistical estimation in high dimensions (i.e. when the dimensionality of the feature space approaches the number of samples) where the maximum likelihood estimator for $\beta$ in \eqref{eq:logregML} is ill-defined. This becomes a problem when the set of OSFs becomes large. We tackle the budgeted detection problem and the high-dimensional estimation problem in a unified manner by considering the set of $\gamma-$sparse models $\{ \beta : \Vert\beta\Vert_0 \leq \gamma \}$ where $\gamma$ is a positive integer and $\Vert \beta \Vert_0$ is the $\ell_0$ ``norm'' which counts the number of non-zero elements in $\beta$. 

We enforce this condition by updating our estimate of $\beta$ in a coordinate-wise manner using \emph{Orthogonal Matching Pursuit} (OMP) as described in Algorithm \ref{alg:OMP}. OMP for logistic regression is an iterative algorithm which starts with a candidate solution $\beta^0$ consisting of the zero vector \cite{lozano2011group}. At each step $t$ it adds a single non-zero coordinate to the solution. This coordinate $j$ is selected according to the criterion in line 2 and is added to the set of selected coordinates, $\mathcal{A} \leftarrow \{j\} \cup \mathcal{A}$. $R^{t} = y -  \sigma({\beta_{\mathcal{A}}^{t}} \tr \Phi_{\mathcal{A}})$ is the vector of residual errors at iteration $t$. The model is then updated by solving the logistic regression problem as in line 4 
where the subscript ${\mathcal{A}}$ considers only the coordinates indexed in $\mathcal{A}$ and all others remain zero. 

These steps are repeated for $t=1,\ldots,\gamma$, that is until there are at most $\gamma$ non-zero elements in $\beta$.


Recently, a variant of OMP, budgeted OMP, which takes feature cost into account has been proposed \cite{grubb2014anytime}. The only difference is to line 2 of Algorithm \ref{alg:OMP}. Given an active set $\mathcal{A}$ of selected features, the next feature is instead selected according to:
$$
j = \arg\max_{j\notin\mathcal{A}} \frac{| \Phi_j\tr r | }{c(j) \cdot \Phi_j\tr\Phi_j}.
$$
This introduces a trade-off between the utility of a particular feature (in terms of reducing the residual training error) and the cost of that feature. As such when $|\mathcal{A}|=d$, we also obtain an ordering of the features according to this trade-off. This allows predictions to be made at test-time to fit a particular computational budget $C$ by selecting the top $g$ features such that $\sum_{j=1}^g c(\mathcal{A}_j) \leq C$, where $\mathcal{A}_j$ is the $j^{th}$ element added to the active set. \cite{grubb2014anytime} show that for a given budget, budgeted OMP returns a solution which is close to optimal. 

\subsection{Stability Selection}

As explained in Section \ref{sec:bag}, since the outlier detection problem is highly imbalanced, bagging with non-uniform sampling is necessary to learn a good classifier. This presents the problem that due to random fluctuations introduced in the sub-sampling step, the order in which features are selected by budgeted OMP in each bag may be different. The question of which features to include in the final model is answered by \emph{stability selection} \cite{Meinshausen2010}. Stability selection is a model selection based on comparing multiple models trained on subsamples of the data. Under certain conditions for a linear model (and assuming uniform $n/2$ bootstrap sampling), it has been shown that the set of features returned by stability selection for OMP correspond to the true underlying model \cite{Meinshausen2010}. In order to deal with our budget constraint, we present a slightly modified version of stability selection.

For each bag $b=1,\ldots,B$ we obtain the solution vector which satisfies a given budget constraint $C$ using budgeted OMP which we denote as $\beta^{(b)}$. 
The set of selected features is then $\mathcal{A}^{(b)} = \{j: \beta^{(b)}_{j} \neq 0, ~~ j=1,\ldots,d \}$. For each feature $j$ we count the proportion of times it was selected across the bags as $J_j = \frac{1}{B} \sum_{b=1}^B \mathbb{I}\{j\in\mathcal{A}^{(b)}\}$, where $\mathbb{I}$ is the binary indicator function.

Now, in order to ensure that the constraint is satisfied in the final model, we construct a \emph{stable set} of features as the set of most commonly selected features across all bags which satisfy the budget constraint. Denoting $\Pi(J)$ as the permutation which sorts the elements of $J$ in descending order, the stable set is then
$$
\mathcal{S}_C = \{ j: \sum_{j\in \Pi(J)} c(j) \leq C \}.
$$ 
That is, we add features to $\mathcal{S}_C$ in order corresponding to how often they were selected across the $B$ bags, until their combined cost matches the specified budget $C$.

Finally, a classifier in each bag is trained using the set of features indexed by $\mathcal{S}_C$. Although the theoretical guarantees about the final model no longer hold due to our imbalanced sampling procedure, we find that our modified stability selection procedure performs well empirically.  

\begin{figure*}[!t]
\begin{center}
\begin{subfigure}[b]{0.32\textwidth}	
\includegraphics[width=1\textwidth]{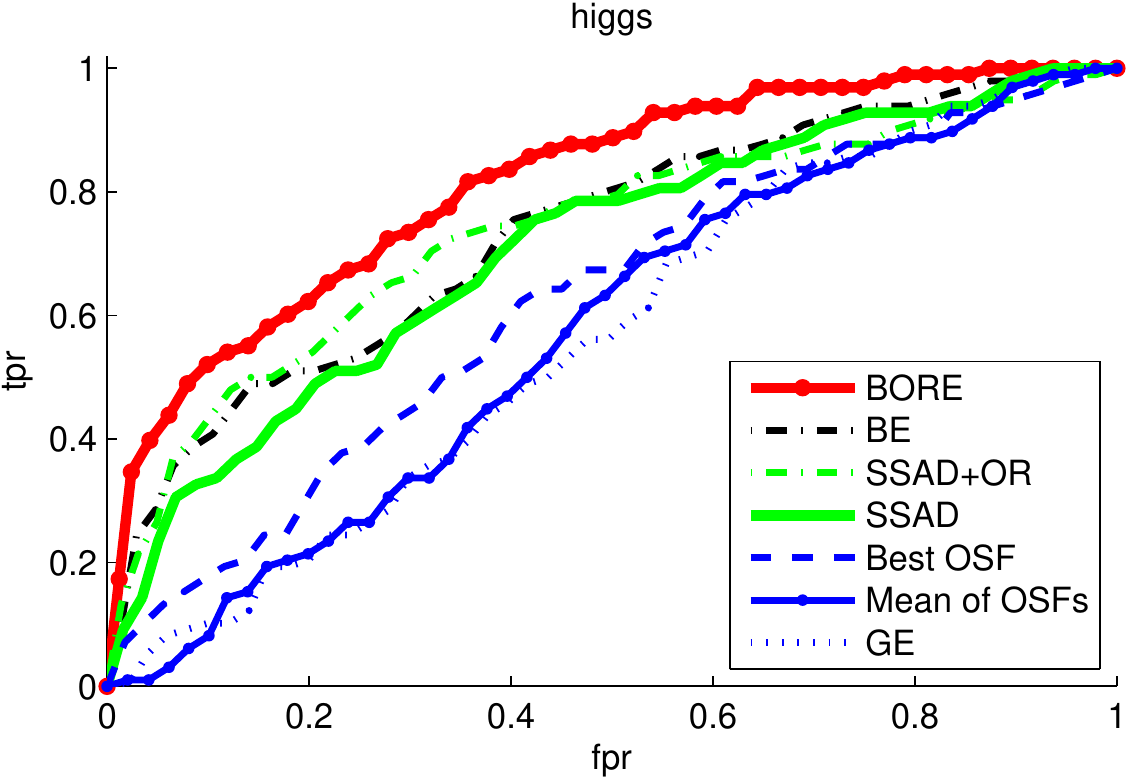}
\end{subfigure} 
\begin{subfigure}[b]{0.32\textwidth}	
\includegraphics[width=1\textwidth]{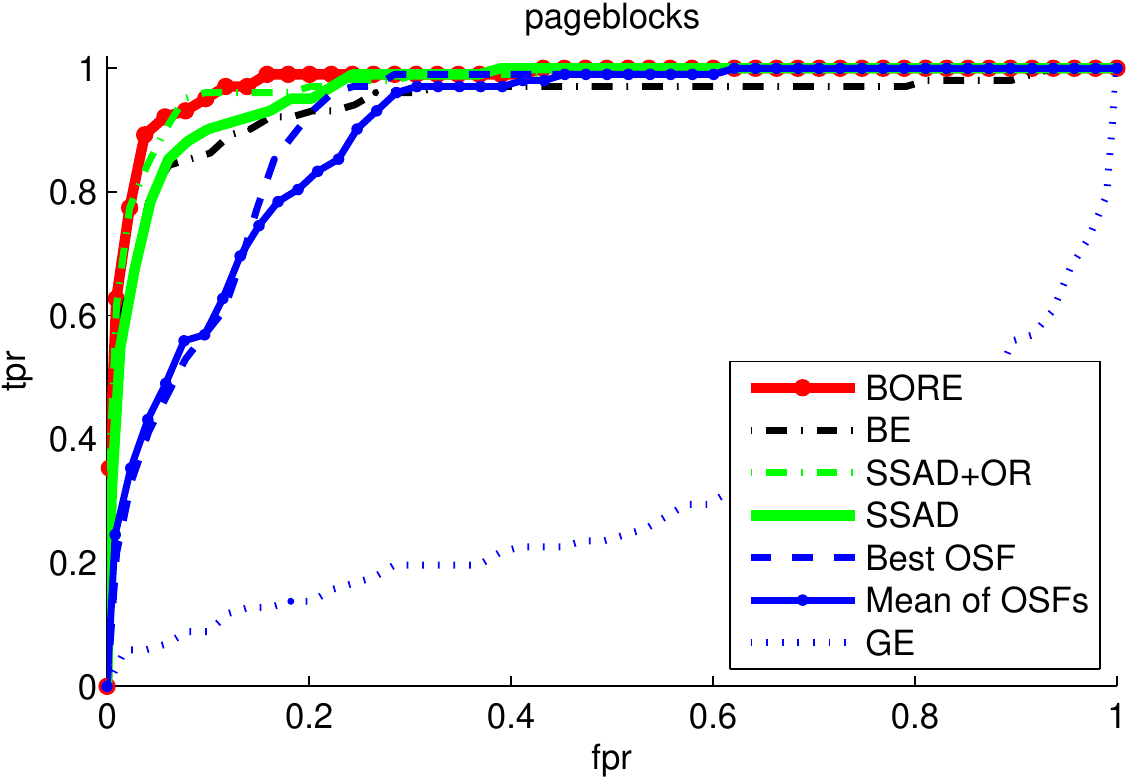}
\end{subfigure} 
\begin{subfigure}[b]{0.32\textwidth}	
\includegraphics[width=1\textwidth]{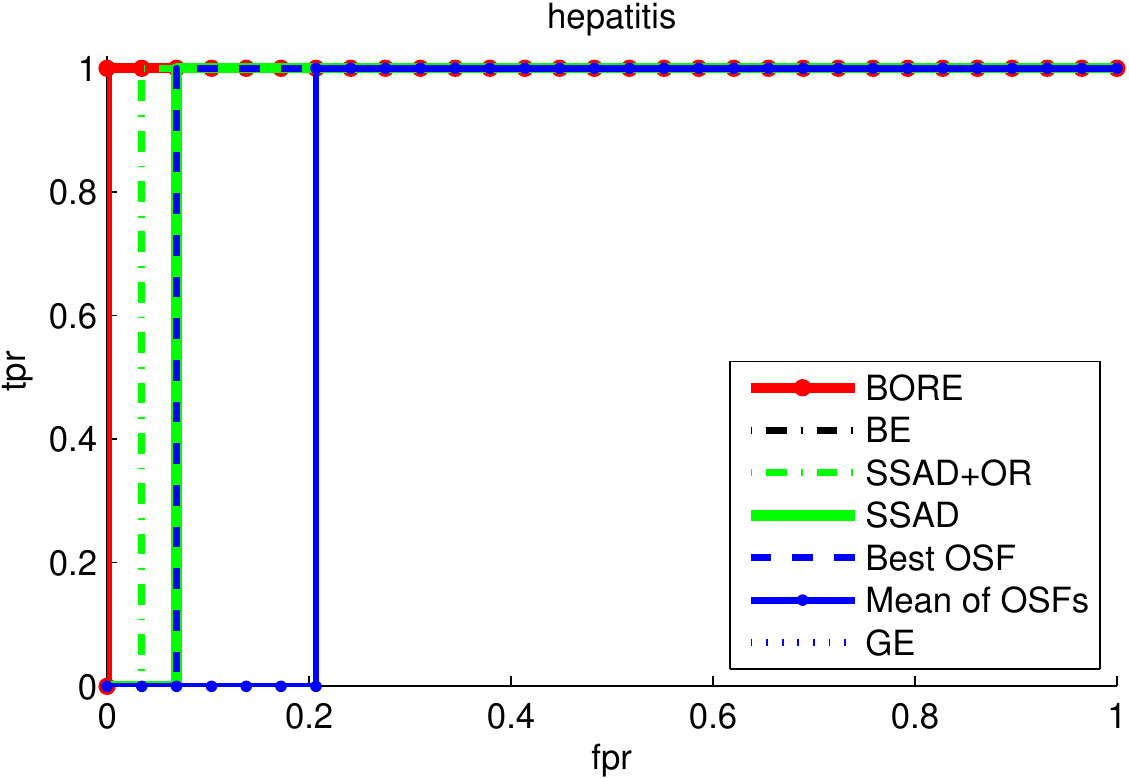}
\end{subfigure} 

\begin{subfigure}[b]{0.32\textwidth}	
\includegraphics[width=1\textwidth]{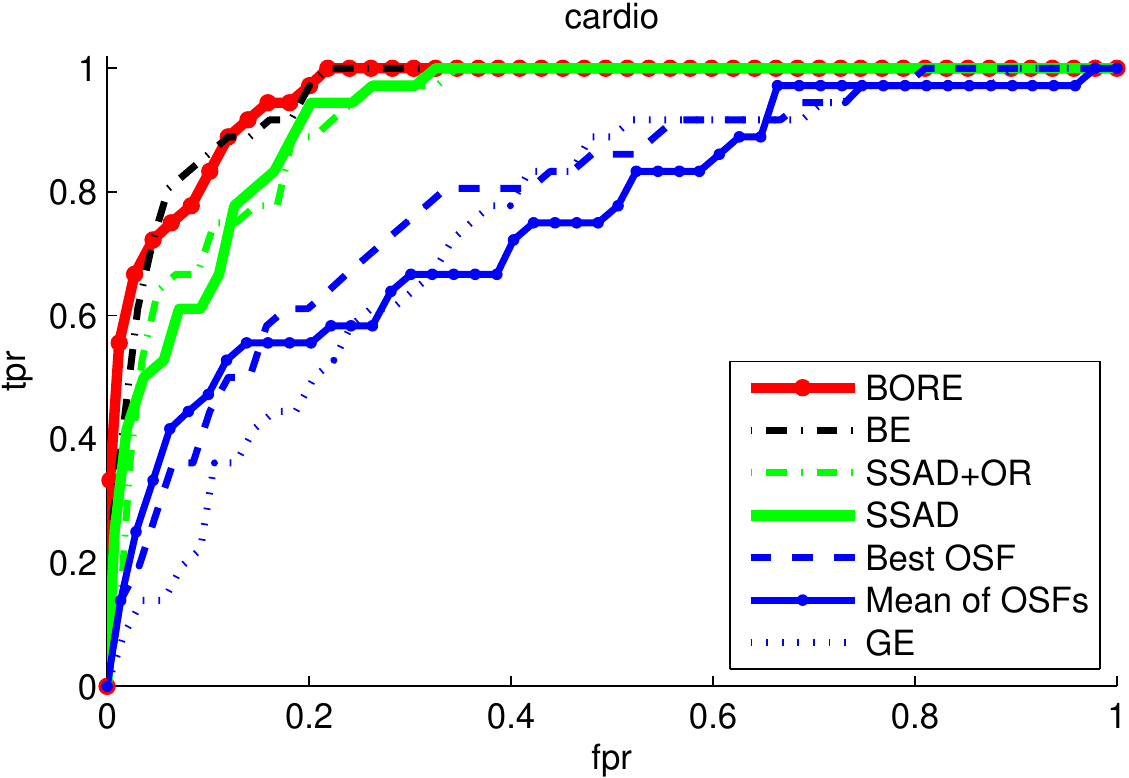}
\end{subfigure} 
\begin{subfigure}[b]{0.32\textwidth}	
\includegraphics[width=1\textwidth]{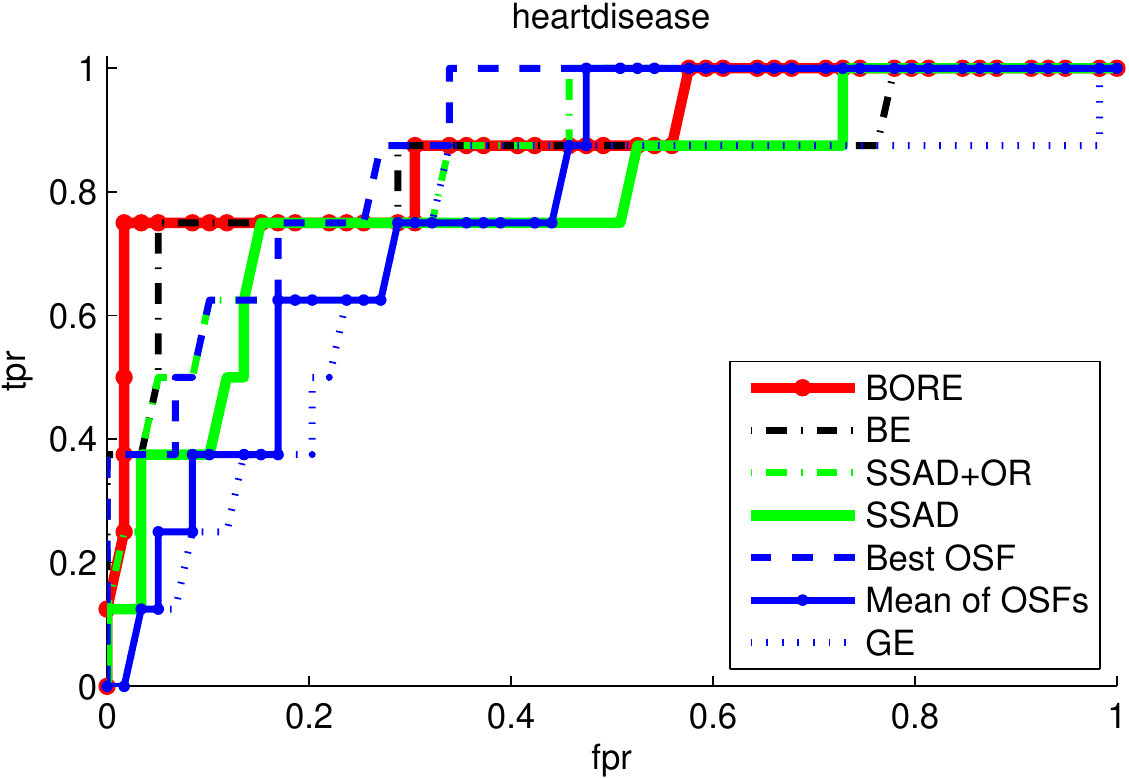}
\end{subfigure} 
\begin{subfigure}[b]{0.32\textwidth}	
\includegraphics[width=1\textwidth]{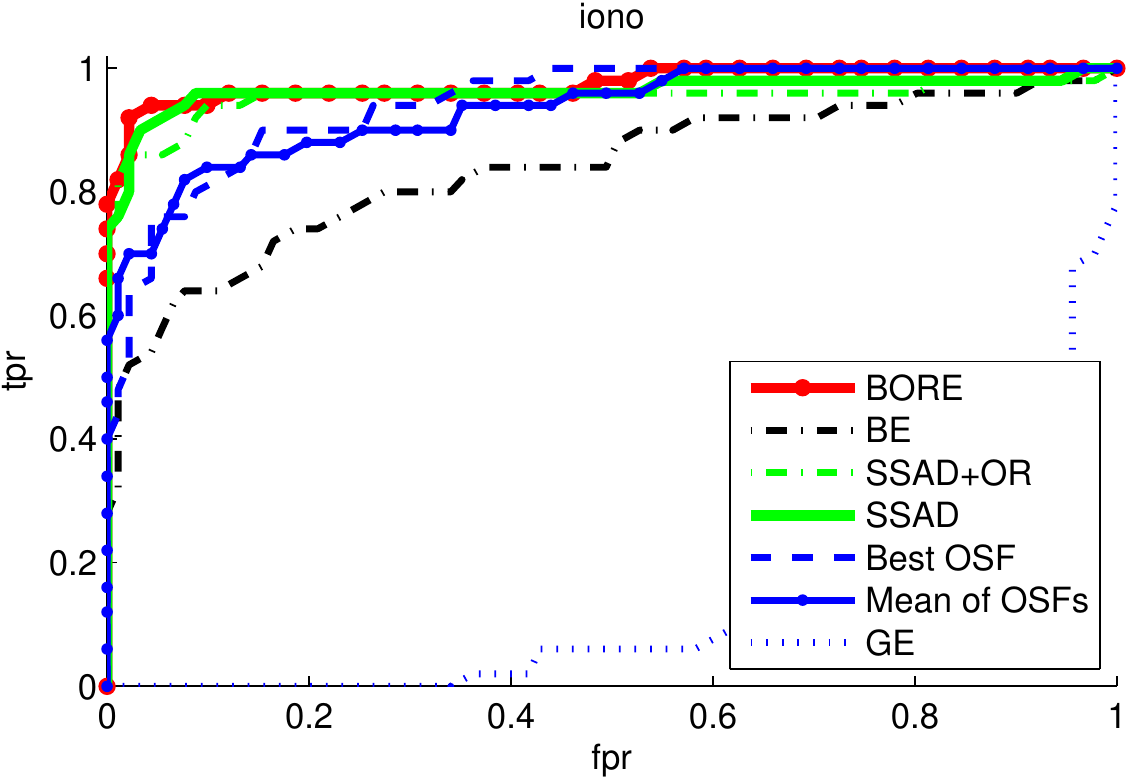}
\end{subfigure} 
\end{center}
\vspace{-0.5cm}
\caption{ROC curves comparing the performance of BORE against 6 competitors on a selection of datasets.}
\vspace{-0.25cm}
\label{fig-bore-roc}

\end{figure*}

\section{Experimental Evaluation} \label{sec-exp}


%

\subsection{Datasets and Features}

We evaluate the performance of BORE on 12 real world datasets summarised in Table \ref{tab:datasets}. The dataset {Higgs} consists of the training set of the Higgs Boson Machine Learning Challenge\footnote{\href{https://www.kaggle.com/c/higgs-boson}{https://www.kaggle.com/c/higgs-boson}} where the goal was to distinguish measurements of Higgs boson particles from the background. We subsampled Higgs bosons such that they form a minority class. The rest of the datasets come from UCI \cite{uci} and they were also preprocessed (by subsampling one or multiple classes) such that they are suitable for the outlier detection task\footnote{Datasets and detailed information about preprocessing available from \href{http://www.dbs.ifi.lmu.de/research/outlier-evaluation}{http://www.dbs.ifi.lmu.de/research/outlier-evaluation}.}. We split each dataset into $60\%$ training and $40\%$ testing data.

{\bf Outlier Scoring Functions.} We use a range of distance and density-based OSFs as our feature transformations. Precisely, it is kNN \cite{RamRasShi00}, kNN-weight \cite{AngPiz05}, ODIN \cite{HauKaeFra04}, LOF \cite{BreKriNgSan00}, SimplifiedLOF \cite{SchZimKri14}, COF \cite{TanCheFuChe02}, INFLO \cite{JinTunHanWan06}, LoOP \cite{KriKroSchZim09a}, LDOF \cite{ZhaHutJin09}, LDF \cite{LatLazPok07}, KDEOS \cite{SchZimKri14a} and FastABOD \cite{KriSchZim08}.\footnote{Any other OSF (fitting Def.\,\ref{def:osf}) could also be used.} Each OSF depends on a neighbourhood parameter which we set as $k\in\{1, 10, 20, \dots ,100\}$ (or less for smaller data sets). Each value of $k$ results in a distinct OSF. In total we obtain 71-132 transformed features per data set. 

For two data sets, {Letter} and {Higgs}, we alternatively use kNN and LOF combined with feature bagging \cite{LazKum05}. That is, we compute the OSFs in different subspaces of the original domain, resulting in 50 and 40 transformed features for these datasets, respectively. 

We begin with a visual analysis of the learned outlier representations (ORs).
Figure \ref{fig:tsne_letter} compares the original features (top row) and the OR (bottom row) for three datasets. For the visualisation we applied t-distributed Stochastic Neighbourhood Embedding (t-SNE) \cite{van2008visualizing} to bags of points subsampled according to the procedure described in Section \ref{sec:bag}. t-SNE finds a low-dimensional non-linear embedding of high-dimensional data which groups similar points together and enforces greater distances between dissimilar points. The visualisations reveal that the learned ORs provide a better separation between the outlying points and the non-outlying points. In contrast, in the original feature space the outliers tend to be more uniformly distributed amongst the non-outlying points. This suggests that classifiers trained on the bags consisting of the OR should achieve higher accuracy than those trained on the original features.

\subsection{Learning Outlier Representations}

\subsubsection{Algorithms} 
To demonstrate the effectiveness of BORE, we perform extensive comparisons with a diverse set of state-of-the-art approaches, both supervised and unsupervised. {Greedy ensemble} (GE) is a purely unsupervised technique of combining OSFs \cite{SchWojZimKri12}. This baseline uses all OSFs but no label information. The {Best OSF} baseline is a single OSF that exhibits best performance in hindsight. It should be noted that this cannot be realised in practice since it requires knowing the out-of-sample performance of each method \emph{a priori}. 
{Mean of OSFs} simply averages the output of all OSFs. {SSAD} \cite{jair2013-semisup} is a semi-supervised outlier detection technique using exactly the same label information as BORE. 

To evaluate our bagging procedure for dealing with imbalance and heterogeneity in the outlier class, we also run SSAD on the data augmented by the new outlier representations {SSAD+OR}. To judge the effectiveness of the outlier representation, we compare against {bagged ensemble} (BE) which builds a bagged model in a manner identical to BORE except only in the original feature space (i.e. it does not learn the outlier representations).

\subsubsection{Experimental Setup} 
For both BORE and BE we construct 50 balanced bags from the training set by subsampling $70\%$ of the labeled outliers and an equivalent number of labeled normal points into each bag. We use BORE and BE \emph{without} $\ell_2$ regularization. Empirically, we observe that the bagging procedure performs implicit regularization and the additional $\ell_2$ penalty is chosen as $0$ by cross validation. Importantly, this means that for a fixed number and size of bags, BORE requires no additional tuning parameters. 

For SSAD and SSAD+OR, we require setting the Gaussian kernel width $\sigma$, and a regularization parameter, $\kappa$. We selected the optimal parameters using 5-fold cross validation for $\sigma=[0.005, 5]$ and $\kappa=[0.5, 20]$. SSAD requires the setting of further parameters which control the regularization of outlier and inliers separately as well as unlabeled data. We leave these at their default values. We scale the data between $[0,1]$ for BORE and normalize by standard deviation for SSAD (this provides the best empirical performance). For GE, we initialize the target vector with the ground truth number of outlier candidates---in reality this is typically not available.

\subsubsection{Results} \label{sec:results1}
Figure \ref{fig-bore-roc} shows receiver operator characteristic (ROC) curves comparing BORE with each of the competitors on 6 datasets.
Table \ref{tab-outDet} contains results on all 12 datasets and it compares each of the methods in terms of standard outlier detection measures (as used e.g.\ in  \cite{jair2013-semisup,Agg13}), namely: area under the ROC curve (AUC), area under the false positive rate interval $[0,0.1]$ of the ROC curve (AUC 0.1) and precision@$n_o$ where $n_o$ is the ground-truth number of outliers.

BORE is the best performing method in all three measures for 6 datasets and performs best in at least one of the measures in 9 of the 12 datasets. For the remaining datasets, BORE is always among the best performing methods. The order of the competing methods changes. 

As expected, the methods which use label information outperform the unsupervised methods. Crucially, BORE almost always outperforms BE which implies that the representation learned by combining OSFs is better for outlier detection than using the original features as also suggested by the visualisations in Figure \ref{fig:tsne_letter}.

However, combining our outlier representation with SSAD does not always yield improved performance. This is perhaps due to the additional feature transformation that SSAD performs using the Gaussian kernel. The use of different kernel functions may improve the performance of SSAD+OR but was not explored. Since the OR feature space is already a non-linear transformation of the data, it is not clear which kernel function would be appropriate in combination with the OSFs to improve performance. 
Alternatively, multiple kernel learning could be used to find a good combination of kernels but this would increase the number of tuning parameters. In this regard, BORE is far less sensitive to its hyperparameters (bag size and number of bags, shown below) whereas SSAD is highly sensitive to kernel width and regularization strength.

\begin{figure}
\begin{center}
\begin{subfigure}[b]{0.222\textwidth}	
	\includegraphics[width=1\textwidth]{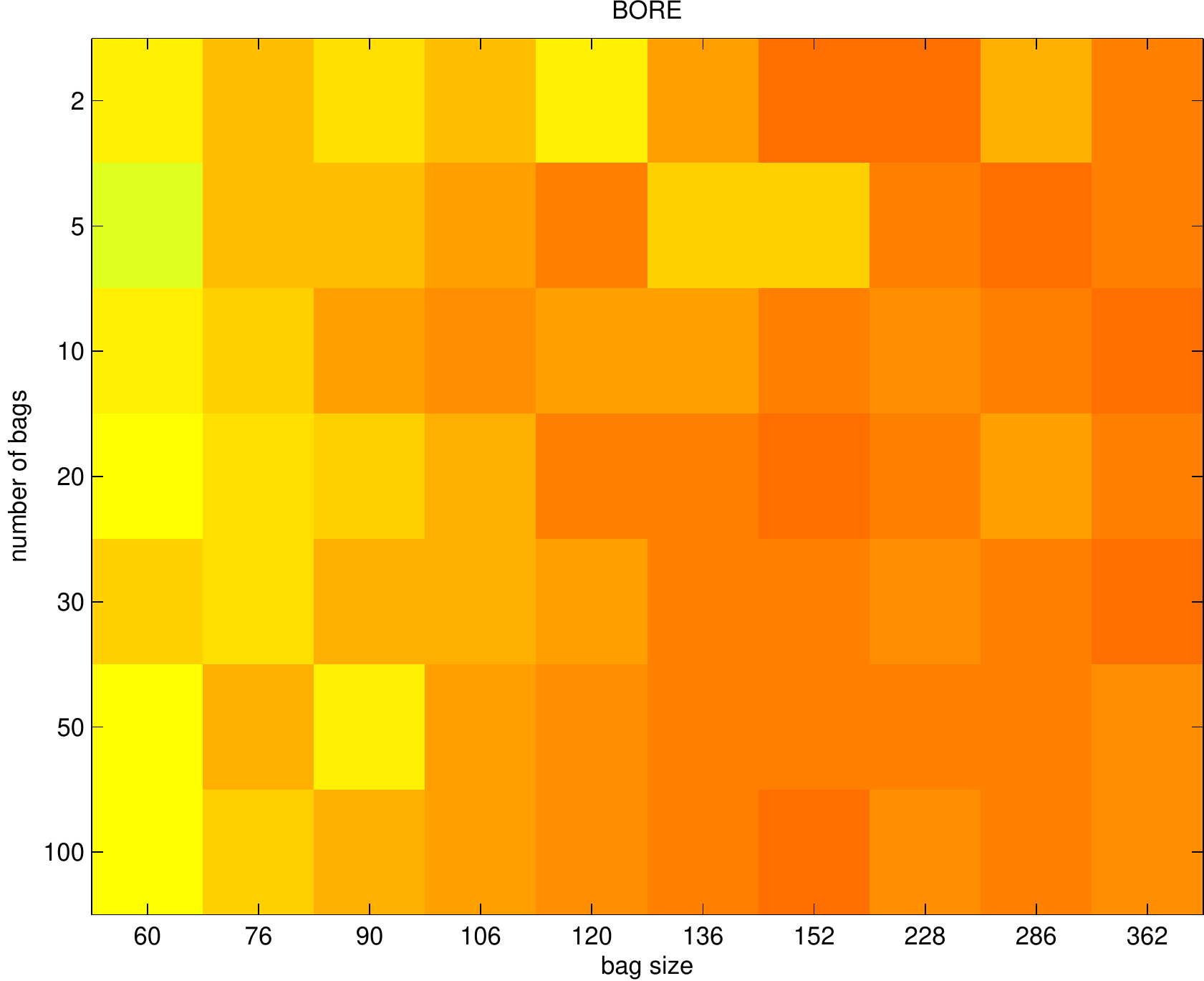}
	\caption{BORE}
\end{subfigure} \begin{subfigure}[b]{0.23\textwidth}		
	\includegraphics[width=1\textwidth]{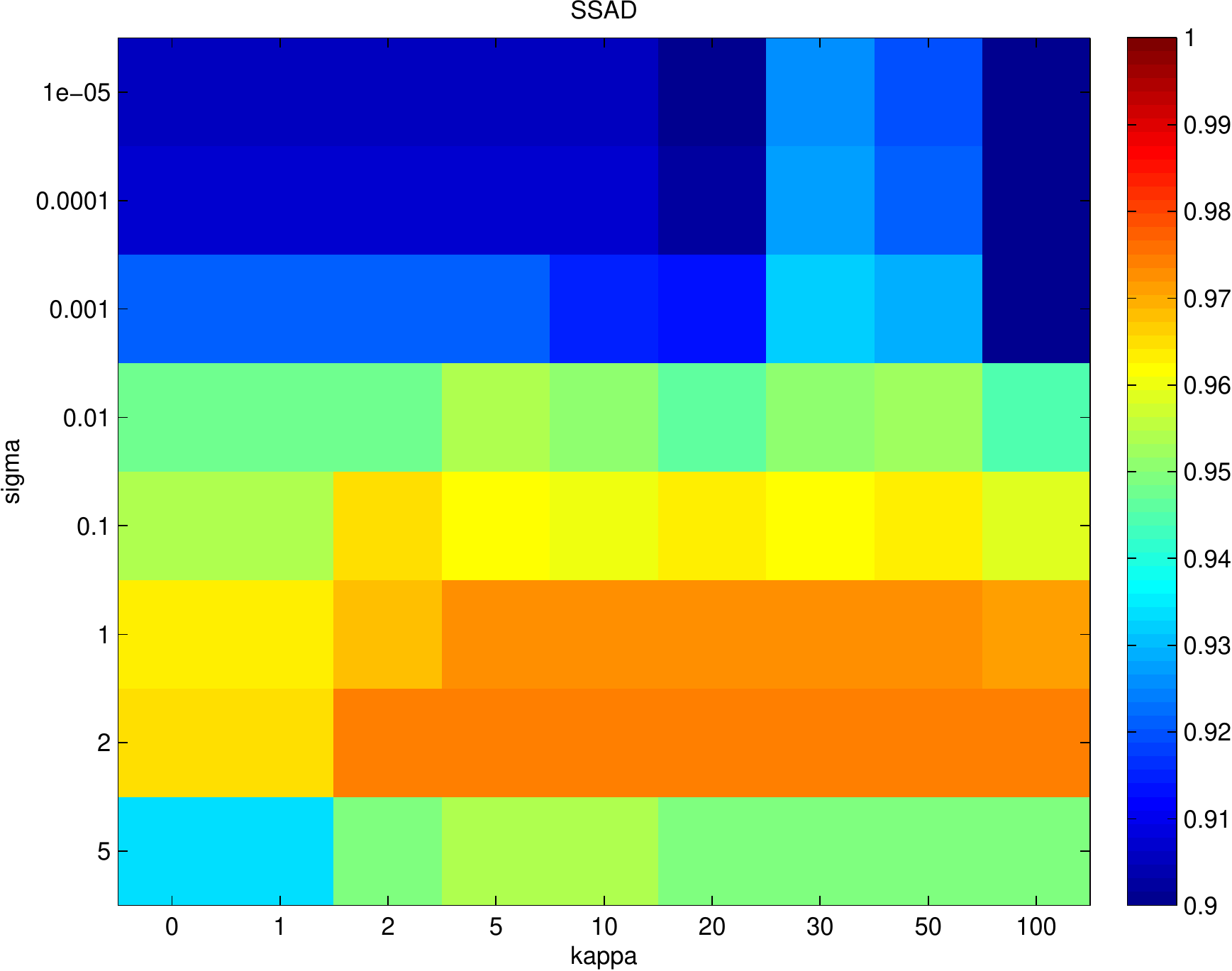}
		\caption{SSAD}
\end{subfigure}
\vspace{-0.3cm}
\caption{Sensitivity of ROC AUC values to different choices of tuning parameters (Ionosphere).}
	\label{fig:bore_sens}
\vspace{-0.7cm}
\end{center}
\end{figure}

Greedy ensemble is often worse than the Mean of OSFs but it only selects a small number of OSFs (between 1 and 13). 
\textbf{Sensitivity to parameters} In Fig.\,\ref{fig:bore_sens}, we compare the sensitivity of BORE and SSAD to their respective parameter settings in terms of AUC. Recall that BORE requires setting on the number of bags and their size whereas SSAD requires a regularization parameter, $\kappa$ and the kernel width, $\sigma$. The AUC achieved by BORE is similar among all parameter values already for a small number of bags. On the other hand, the performance of SSAD varies significantly for different parameters, highlighting the need for cross-validation. 

\begin{table*}[!t]
\begin{center}
\caption{Results of outlier detection for all 12 data sets (in \%). Algorithms marked with $^*$ are unsupervised, others are supervised. \emph{AUC} refers to the area under the ROC curve, \emph{AUC 0.1} is the area under the beginning of the ROC curve (fpr interval [0, 0.1]) and \emph{precision@$n_o$} is precision at the $n_o$-th position in the outlier ranking where $n_o$ is the ground-truth number of outliers in the data set. Bold denotes the best performance for a given evaluation measure and dataset. \label{tab-outDet}
}

\begin{tabular}{ l r r r|r r r|r r r } \hline
&\multicolumn{3}{c|}{cardio}&\multicolumn{3}{c|}{heartdisease}&\multicolumn{3}{c}{hepatitis}\\
&\small{AUC}&\small{AUC 0.1}& \small{precision@$n_o$} & \small{AUC}&\small{AUC 0.1}& \small{precision@$n_o$} & \small{AUC}&\small{AUC 0.1}& \small{precision@$n_o$}\\ \hline
Best OSF$^*$&79.08&26.00&27.78&88.14&40.00&50.00&93.10&0.00&0.00\\
Mean of OSFs$^*$&75.75&31.19&33.33&79.03&15.00&37.50&79.31&0.00&0.00\\
GE$^*$&74.75&13.84&13.89&71.82&7.50&25.00&79.31&0.00&0.00\\
SSAD&92.50&48.66&47.22&78.81&27.50&37.50&93.10&0.00&0.00\\
SSAD+OR&92.49&50.97&52.78&86.44&37.50&50.00&96.55&50.00&0.00\\
BE&95.77&65.72&55.56&84.96&52.50&62.50&93.10&0.00&0.00\\
BORE& \textbf{95.98}& \textbf{66.83}& \textbf{63.89}& \textbf{88.35}& \textbf{65.00}& \textbf{75.00}& \textbf{100.00}& \textbf{100.00}& \textbf{100.00}\\
\hline
\end{tabular}

\vspace{5mm}

\begin{tabular}{ l r r r|r r r|r r r } \hline
&\multicolumn{3}{c|}{higgs}&\multicolumn{3}{c|}{iono}&\multicolumn{3}{c}{letter}\\
&\small{AUC}&\small{AUC 0.1}& \small{precision@$n_o$} & \small{AUC}&\small{AUC 0.1}& \small{precision@$n_o$} & \small{AUC}&\small{AUC 0.1}& \small{precision@$n_o$}\\ \hline
Best OSF$^*$&62.30&11.19&12.24&94.22&68.00&82.00&91.16&44.15&38.89\\
Mean of OSFs$^*$&56.95&2.63&3.06&93.38&73.89&84.00&89.41&35.79&36.11\\
GE$^*$&56.31&5.53&8.16&12.23&0.00&6.00&84.38&19.23&25.00\\
SSAD&70.75&20.98&21.43&96.29&88.13&92.00& \textbf{97.22}&75.41&66.67\\
SSAD+OR&73.55&26.55&25.51&95.41&86.56&88.00&96.22&75.80& \textbf{69.44}\\
BE&73.40&26.55&25.51&83.74&55.11&70.00&84.88&35.09&33.33\\
BORE& \textbf{81.47}& \textbf{37.95}& \textbf{35.71}& \textbf{97.47}& \textbf{91.11}& \textbf{94.00}&96.42& \textbf{76.70}&66.67\\
\hline
\end{tabular}

\vspace{5mm}

\begin{tabular}{ l r r r|r r r|r r r } \hline
&\multicolumn{3}{c|}{pima}&\multicolumn{3}{c|}{pageblocks}&\multicolumn{3}{c}{parkinson}\\
&\small{AUC}&\small{AUC 0.1}& \small{precision@$n_o$} & \small{AUC}&\small{AUC 0.1}& \small{precision@$n_o$} & \small{AUC}&\small{AUC 0.1}& \small{precision@$n_o$}\\ \hline
Best OSF$^*$& \textbf{71.27}&11.13&15.38&91.13&40.61&37.25&91.58&40.00&60.00\\
Mean of OSFs$^*$&64.82&0.81&0.00&90.10&42.96&37.25&83.16&40.00&40.00\\
GE$^*$&59.63&0.00&0.00&29.36&5.90&6.86&85.26&40.00&60.00\\
SSAD&62.53&4.05&0.00&96.46&72.41&62.75&74.74&40.00&60.00\\
SSAD+OR&66.20&6.62&7.69&97.48&81.99& \textbf{69.61}&88.42&20.00&60.00\\
BE&67.50&11.94&15.38&94.23&73.22&62.75&76.84&60.00&60.00\\
BORE&69.99& \textbf{17.81}& \textbf{23.08}& \textbf{97.93}& \textbf{83.28}&68.63& \textbf{91.58}& \textbf{80.00}& \textbf{80.00}\\
\hline
\end{tabular}

\vspace{5mm}

\begin{tabular}{ l r r r|r r r|r r r } \hline
&\multicolumn{3}{c|}{spambase}&\multicolumn{3}{c|}{waveform}&\multicolumn{3}{c}{wilt}\\
&\small{AUC}&\small{AUC 0.1}& \small{precision@$n_o$} & \small{ROC AUC}&\small{AUC 0.1}& \small{precision@$n_o$} & \small{AUC}&\small{AUC 0.1}& \small{precision@$n_o$}\\ \hline
Best OSF$^*$&85.68&42.16&36.84&75.16&33.82&31.11&80.52&20.19&17.76\\
Mean of OSFs$^*$&85.15&44.54&36.84&73.07&33.62&28.89&74.66&3.94&3.74\\
GE$^*$&83.38&43.94&31.58&70.33&22.78&17.78&70.40&0.00&0.00\\
SSAD&96.14& \textbf{72.95}&42.11&91.84& \textbf{59.82}& \textbf{53.33}& \textbf{98.56}&85.78&74.77\\
SSAD+OR& \textbf{96.78}&65.65&21.05&87.73&37.72&28.89&96.17&77.78&72.90\\
BE&94.21&66.59& \textbf{52.63}&88.07&40.85&33.33&98.03&78.26&66.36\\
BORE&95.50&67.12&47.37& \textbf{91.98}&54.77&40.00&98.38& \textbf{87.29}& \textbf{76.64}\\
\hline
\end{tabular}

\end{center}
\end{table*}

\begin{figure*}[!t]
\begin{center}
\begin{subfigure}[b]{0.49\textwidth}	
\includegraphics[width=0.49\textwidth]{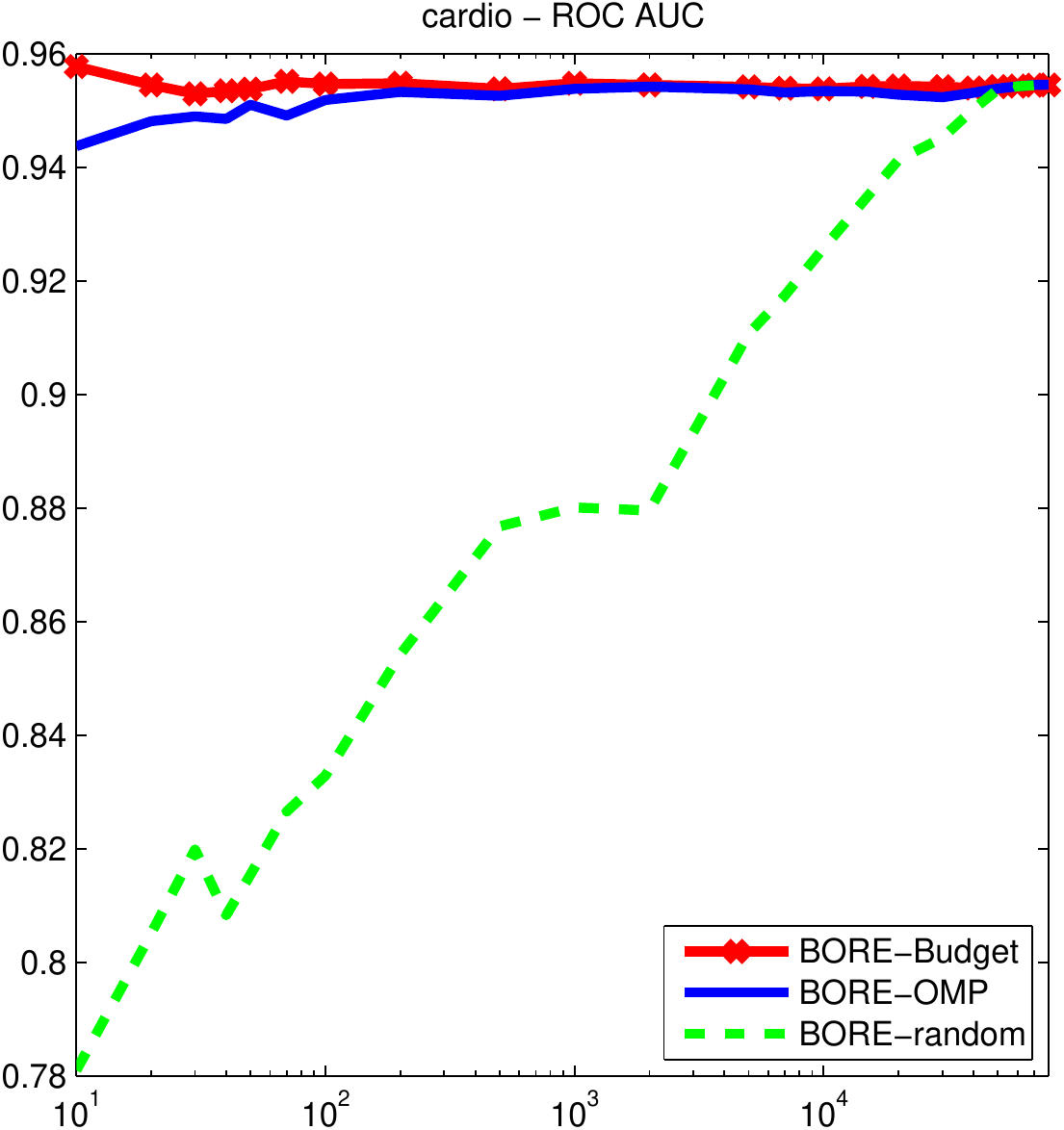}
\includegraphics[width=0.49\textwidth]{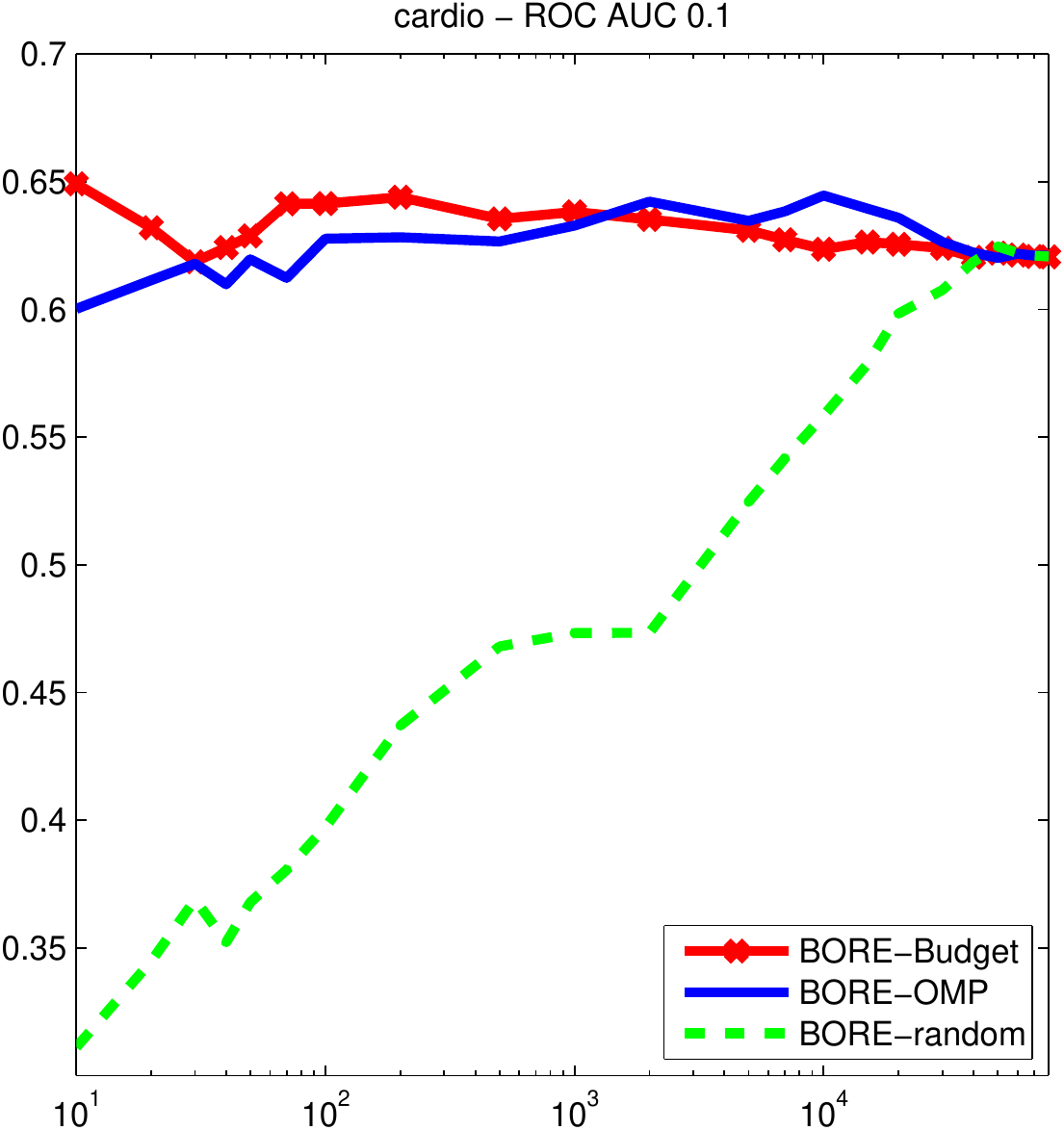}
\subcaption{Cardio.} \label{sub:cardio}
\end{subfigure}
\begin{subfigure}[b]{0.49\textwidth}	
\includegraphics[width=0.49\textwidth]{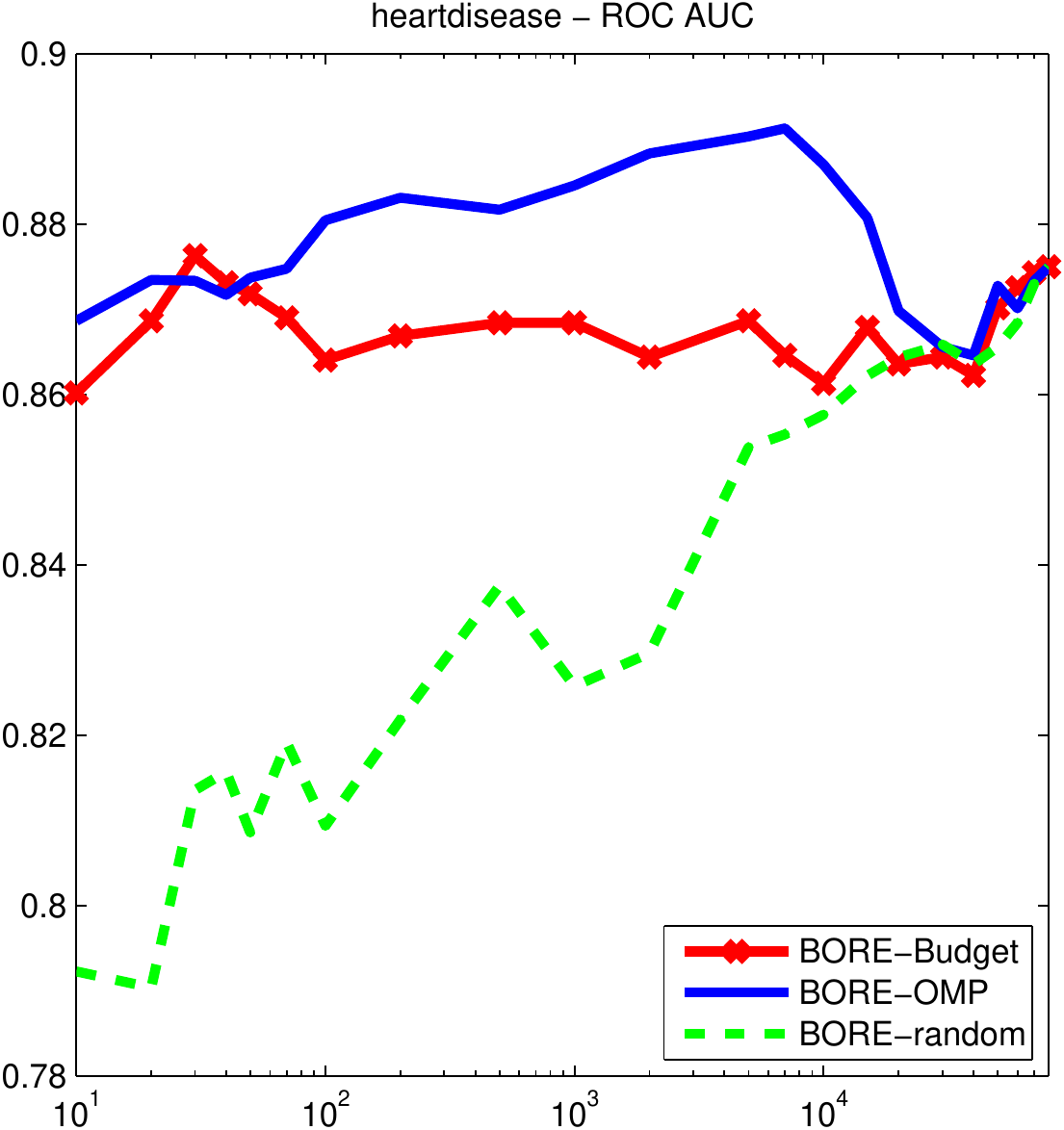}
\includegraphics[width=0.49\textwidth]{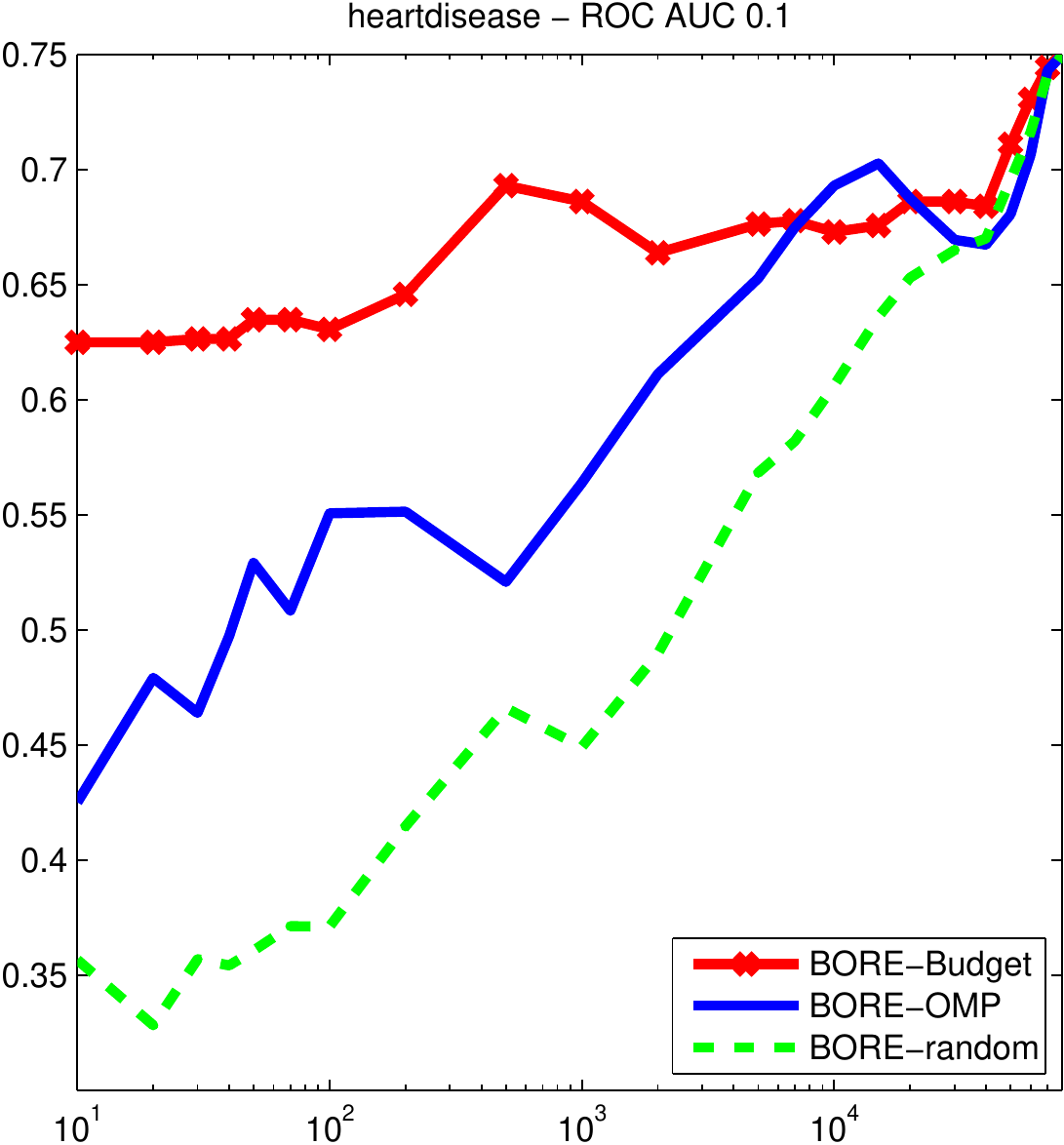}
\subcaption{Heartdisease.} \label{sub:heart}
\end{subfigure}
\begin{subfigure}[b]{0.49\textwidth}	
\includegraphics[width=0.49\textwidth]{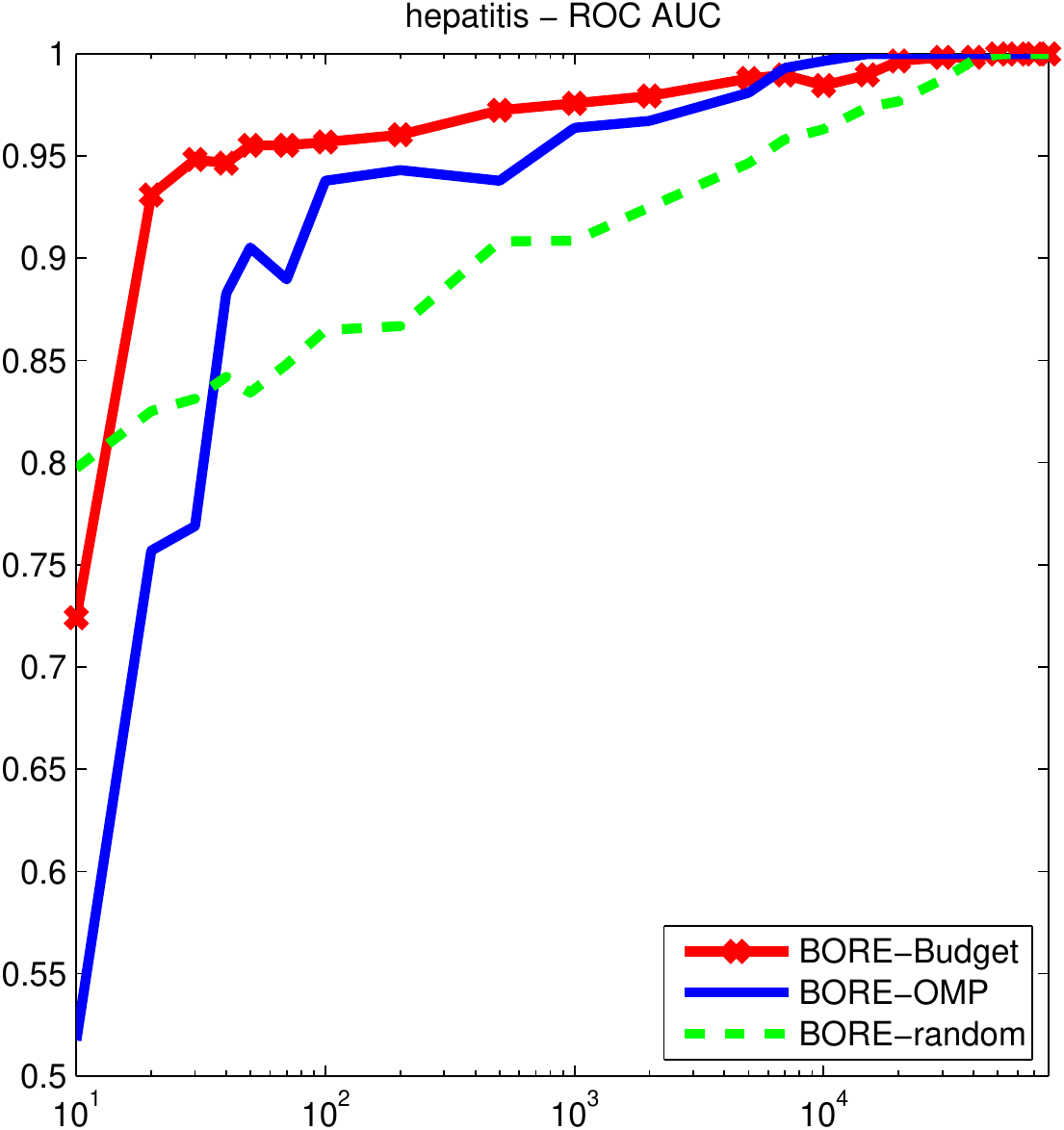}
\includegraphics[width=0.49\textwidth]{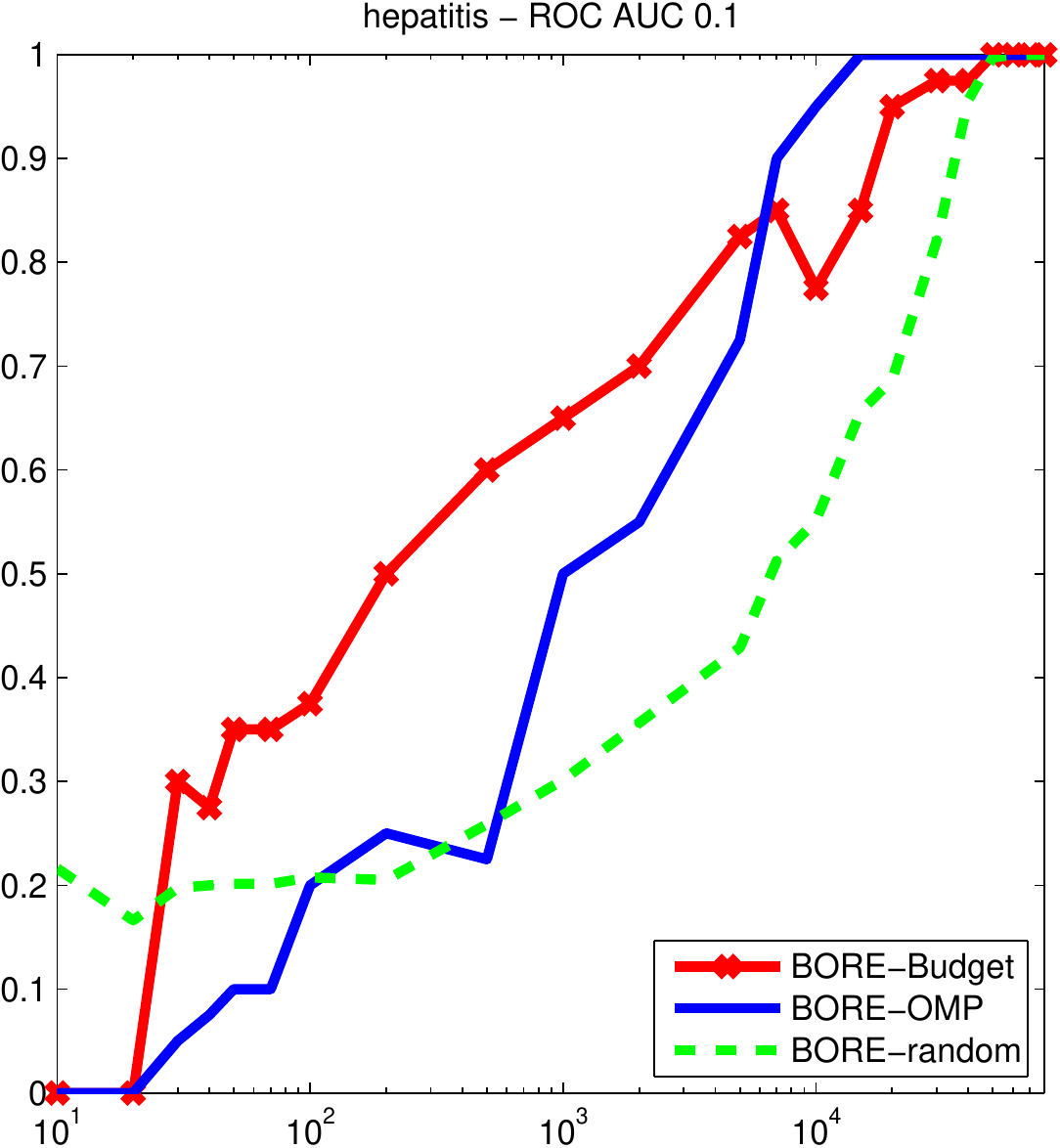}
\subcaption{Hepatitis.} \label{sub:hep}
\end{subfigure}
\begin{subfigure}[b]{0.49\textwidth}	
\includegraphics[width=0.49\textwidth]{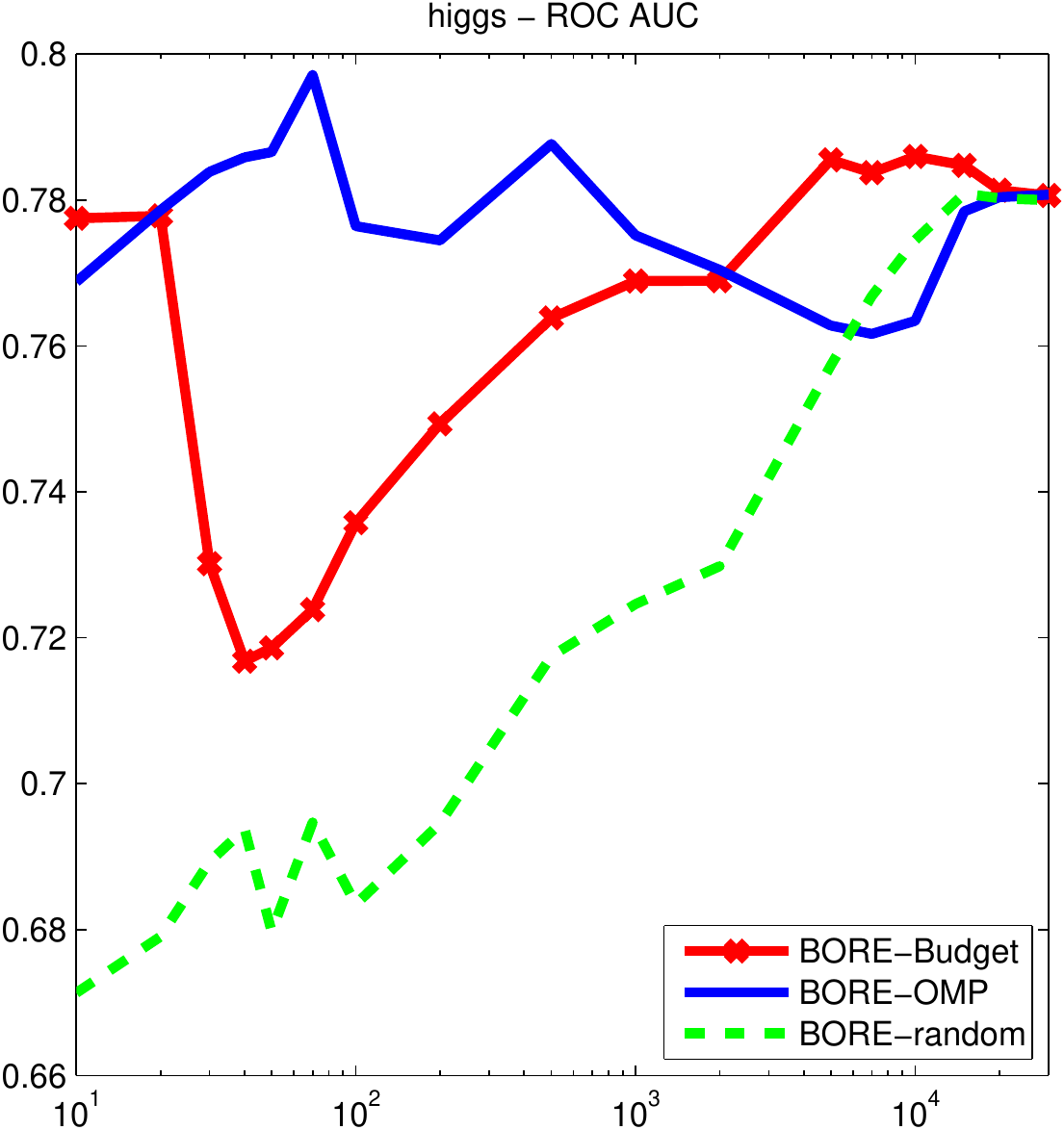}
\includegraphics[width=0.49\textwidth]{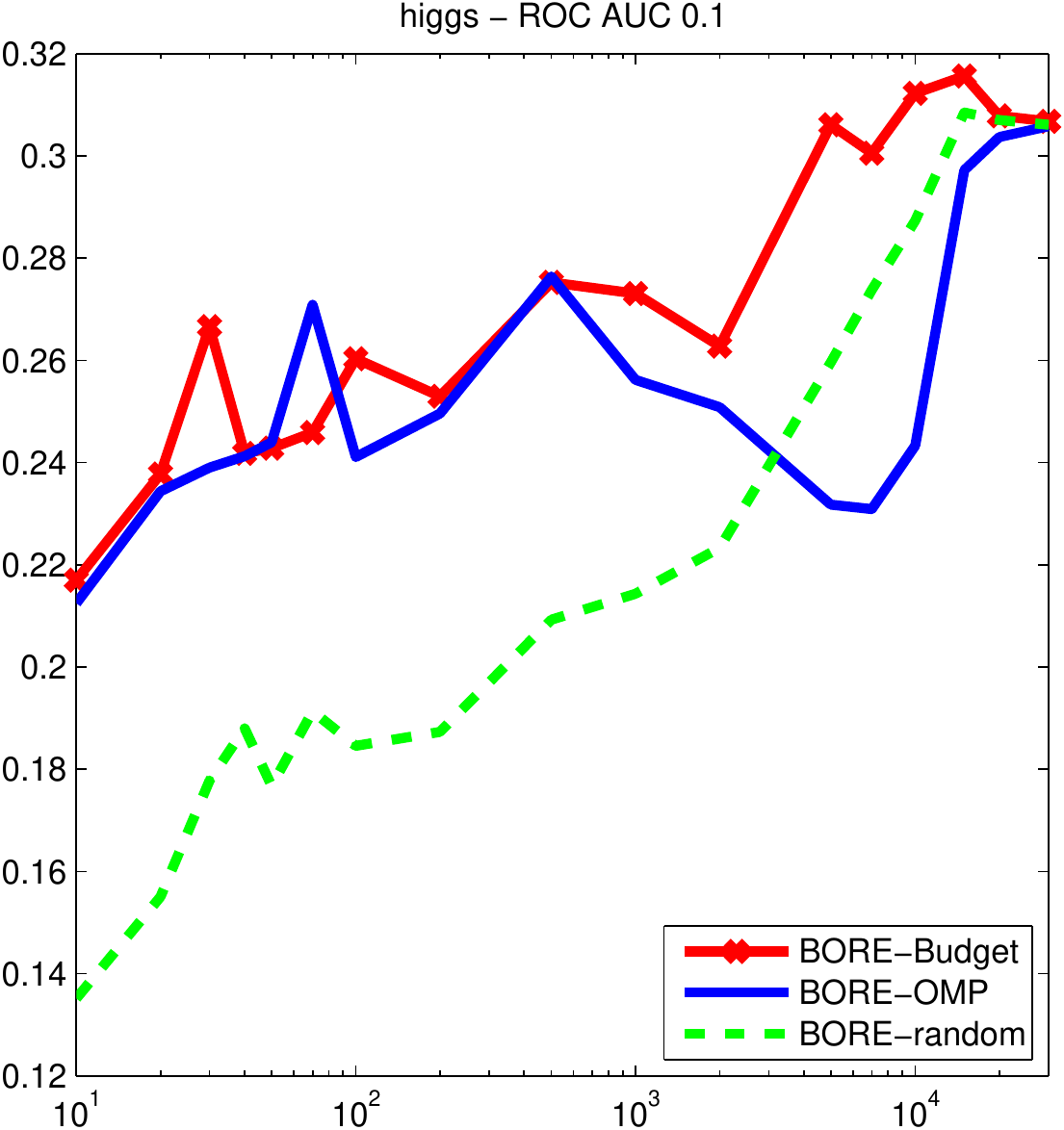}
\subcaption{Higgs.} \label{sub:higgs}
\end{subfigure}
\vspace{-0.3cm}
\caption{Evaluation of outlier detection on budget (first four data sets, see below for remaining data sets). Budgeted OMP is compared to standard OMP and random selection of features. For each data set, we report on ROC AUC and ROC AUC on the false positive rate interval $[0,0.1]$ ($y$-axis) for different budgets ($x$-axis, log-scaled).}
\label{fig:budget_eval1}
\end{center}
\vspace{-1cm}
\end{figure*}

\begin{figure*}[th]
\ContinuedFloat
\begin{center}
\begin{subfigure}[b]{0.49\textwidth}	
\includegraphics[width=0.49\textwidth]{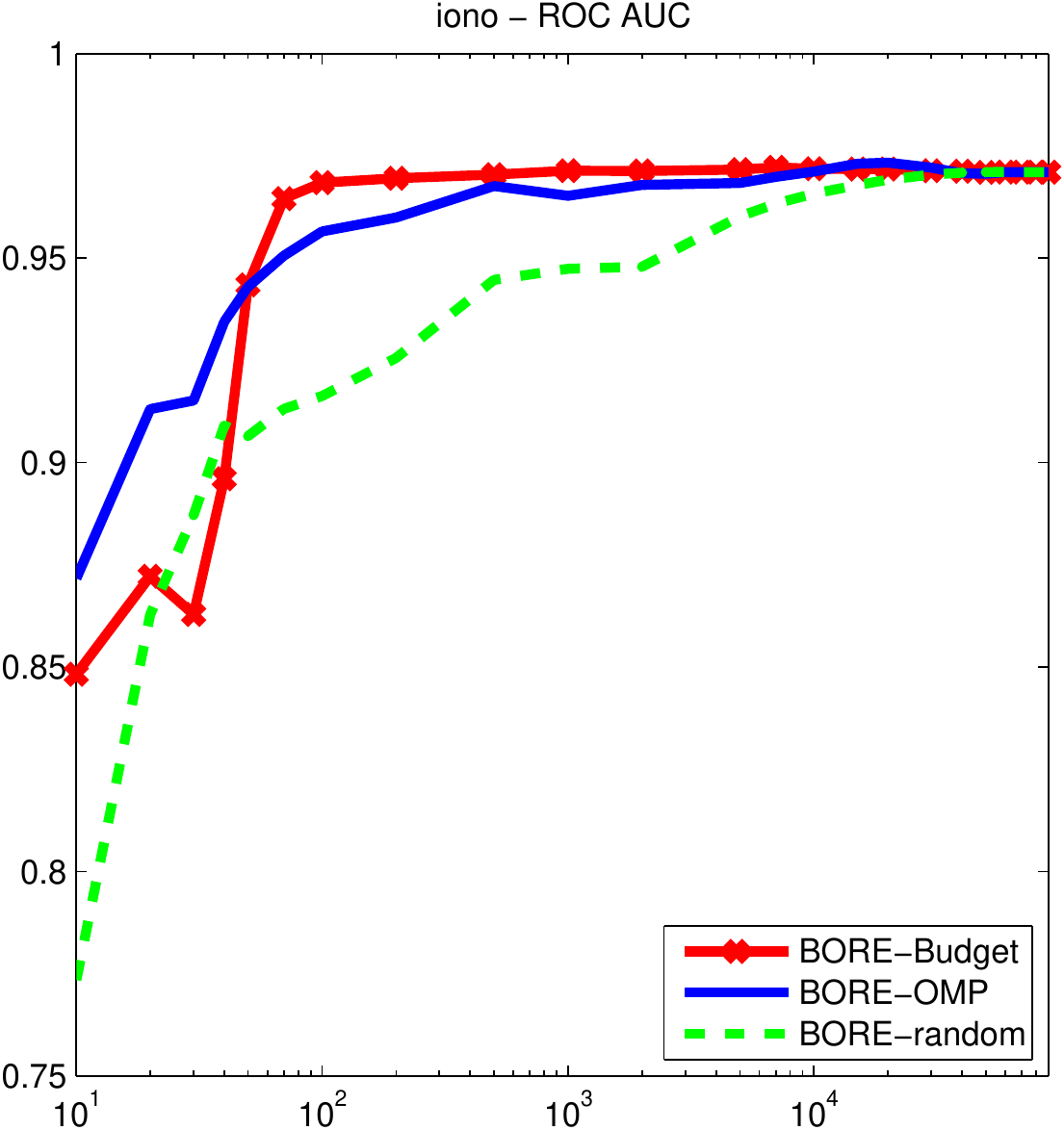}
\includegraphics[width=0.49\textwidth]{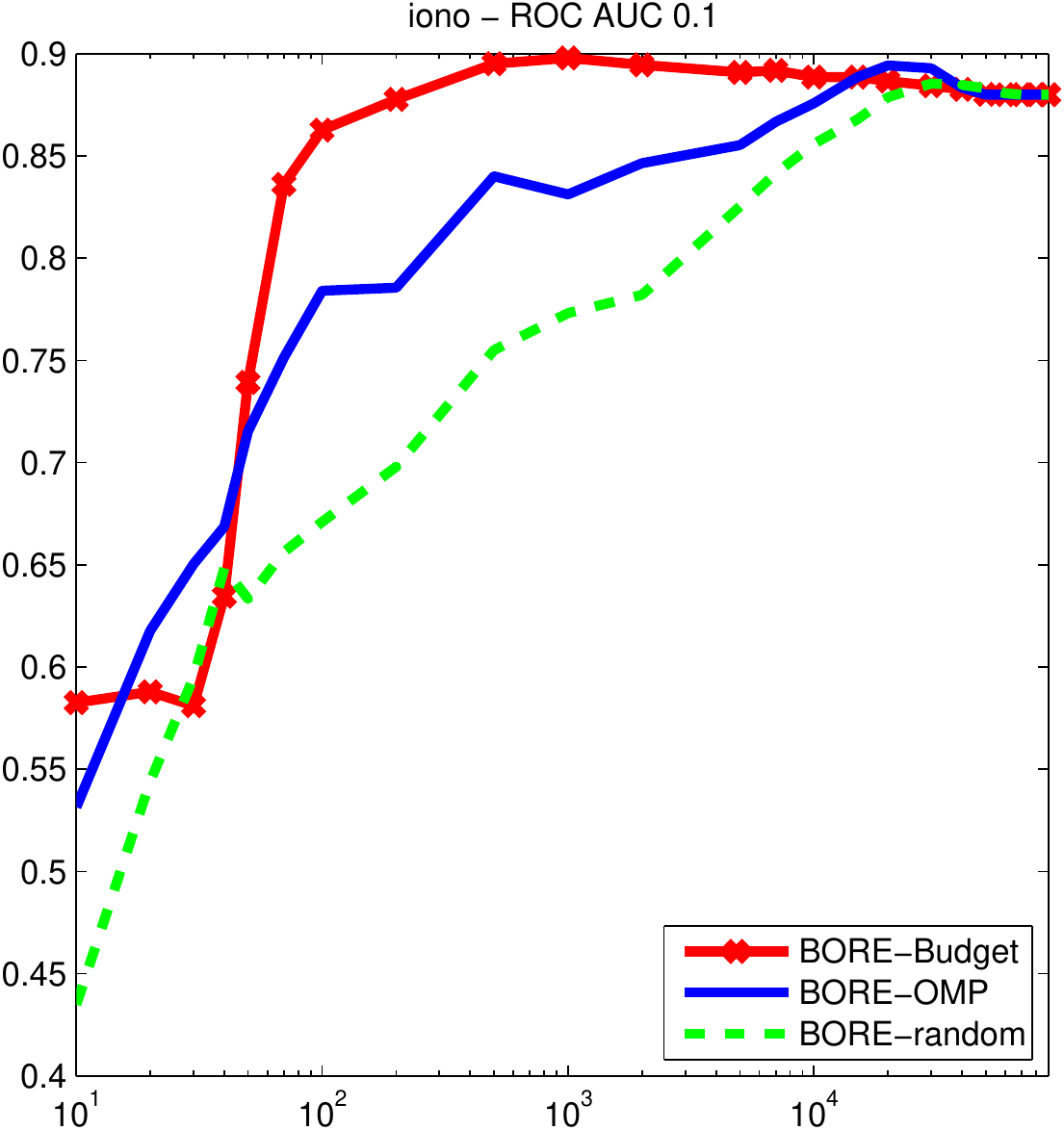}
\subcaption{Ionosphere.} \label{sub:iono}
\end{subfigure}
\begin{subfigure}[b]{0.49\textwidth}	
\includegraphics[width=0.49\textwidth]{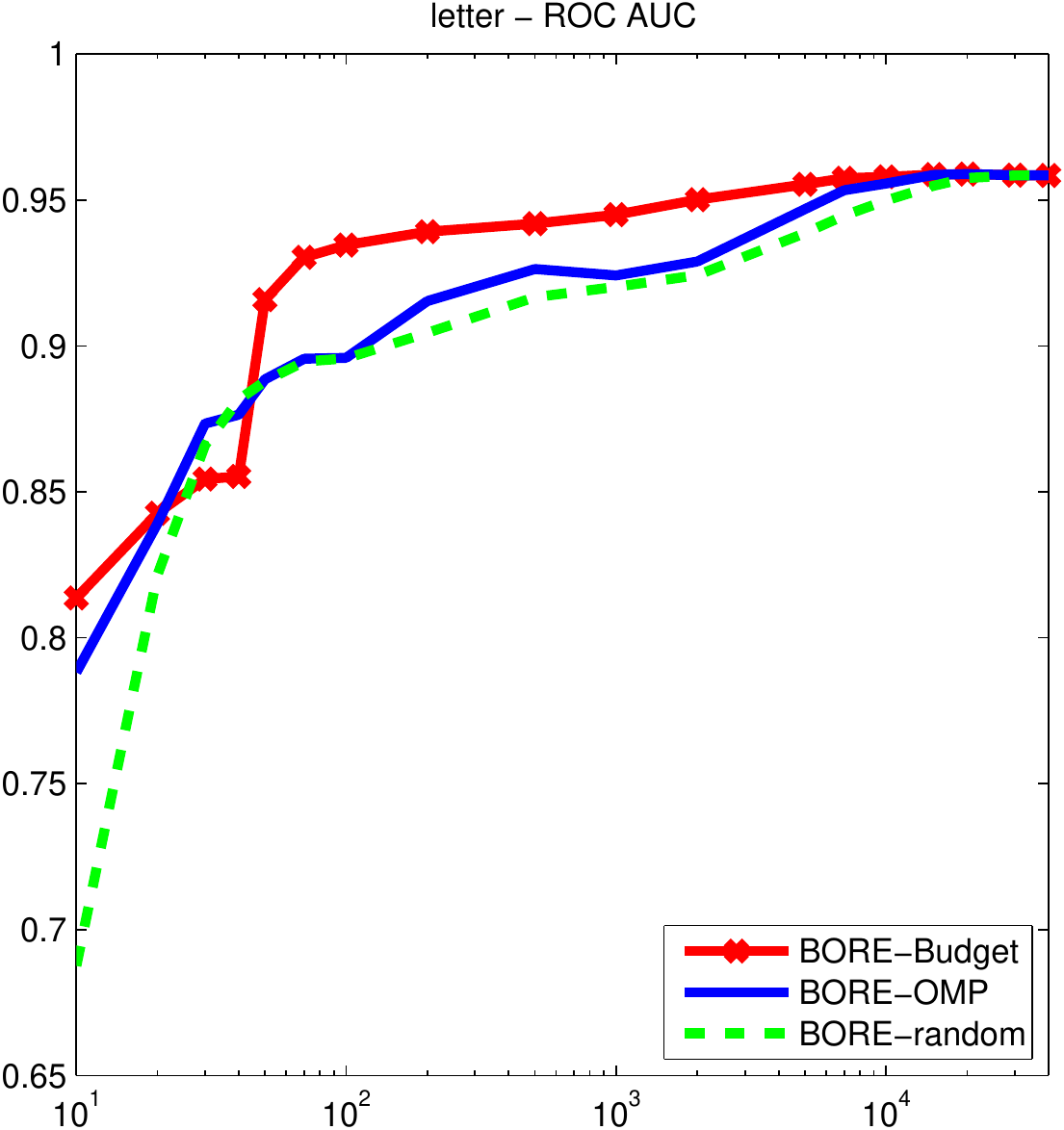}
\includegraphics[width=0.49\textwidth]{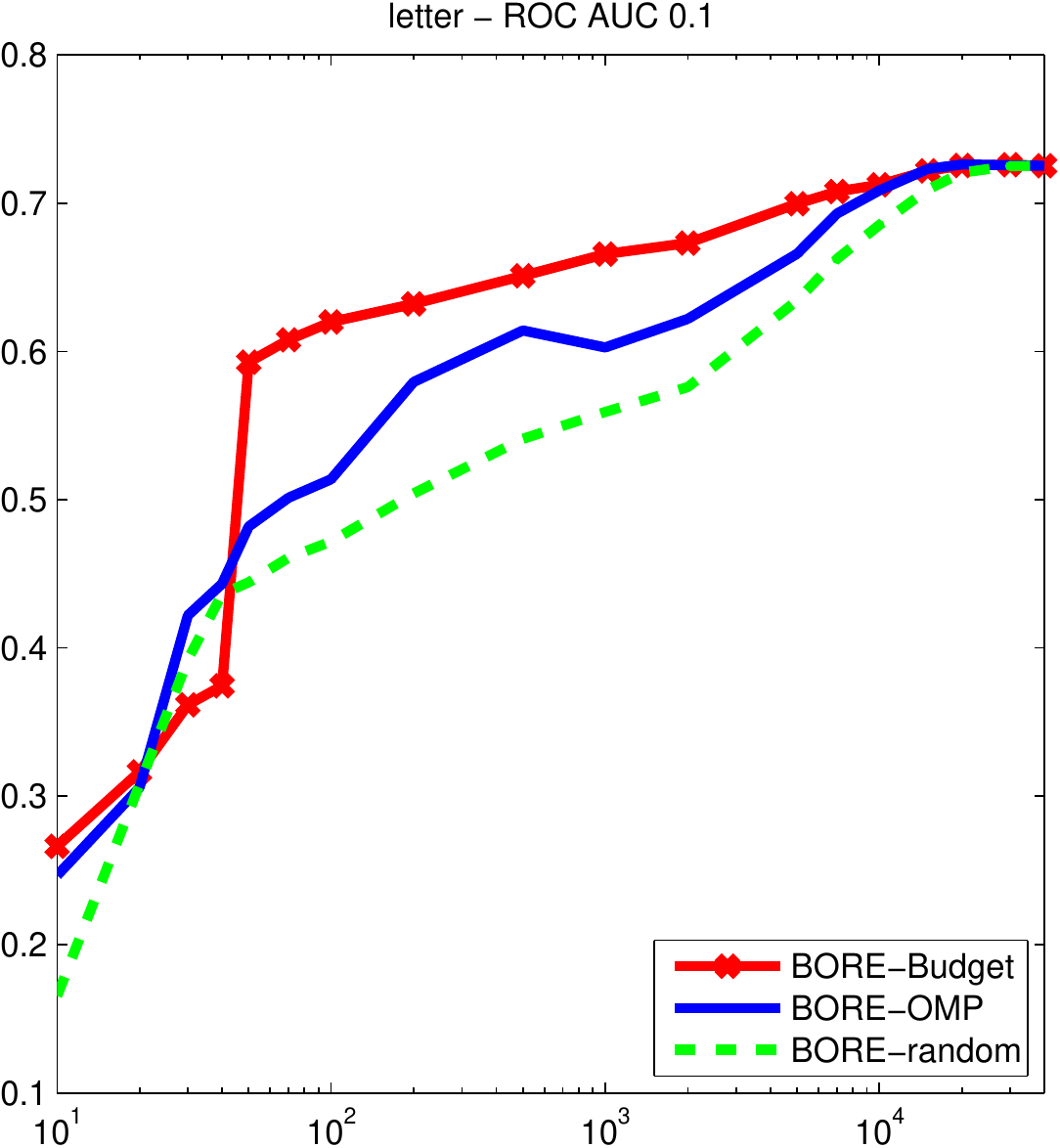}
\subcaption{Letter.} \label{sub:letter}
\end{subfigure}
\begin{subfigure}[b]{0.49\textwidth}	
\includegraphics[width=0.49\textwidth]{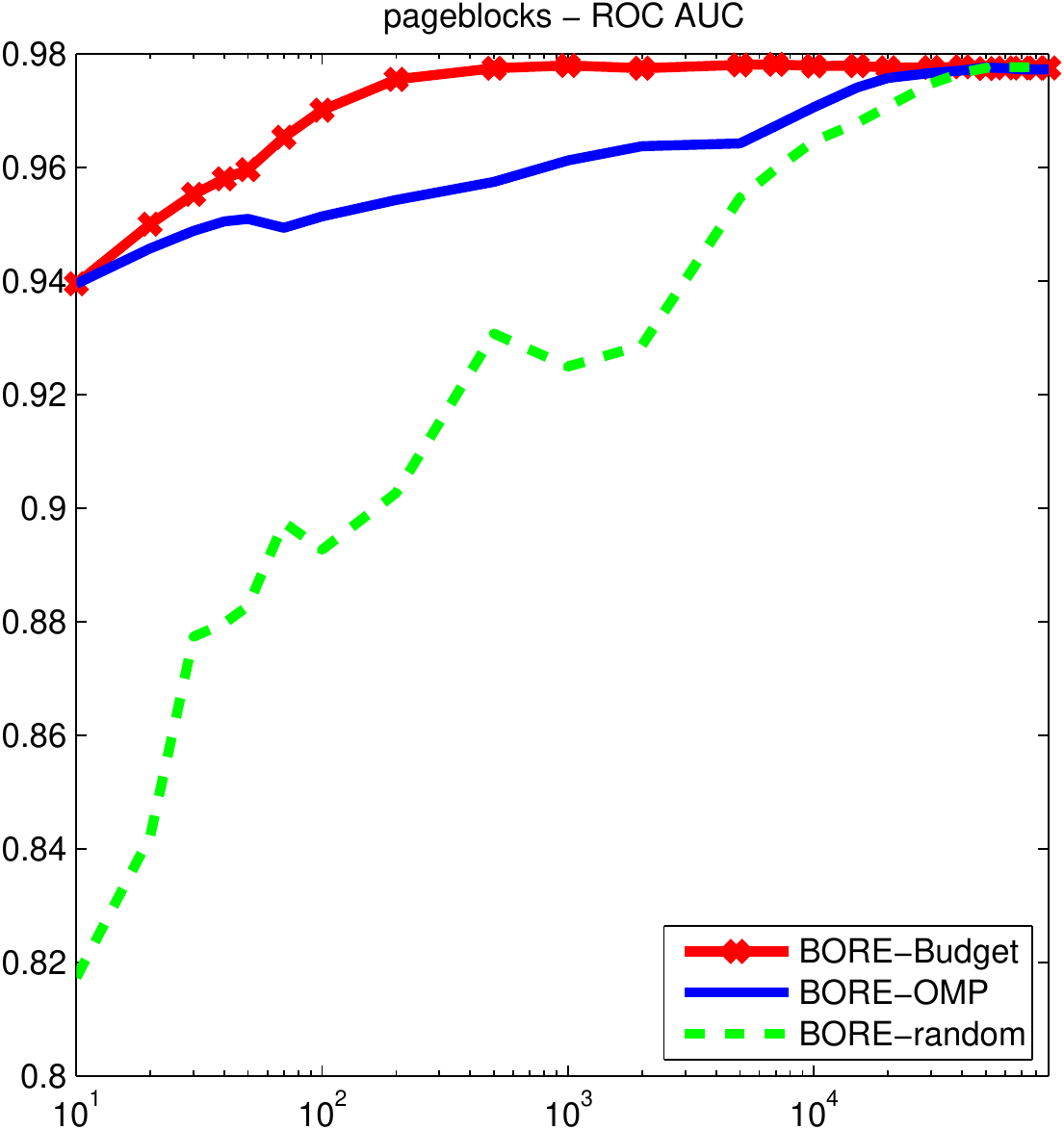}
\includegraphics[width=0.49\textwidth]{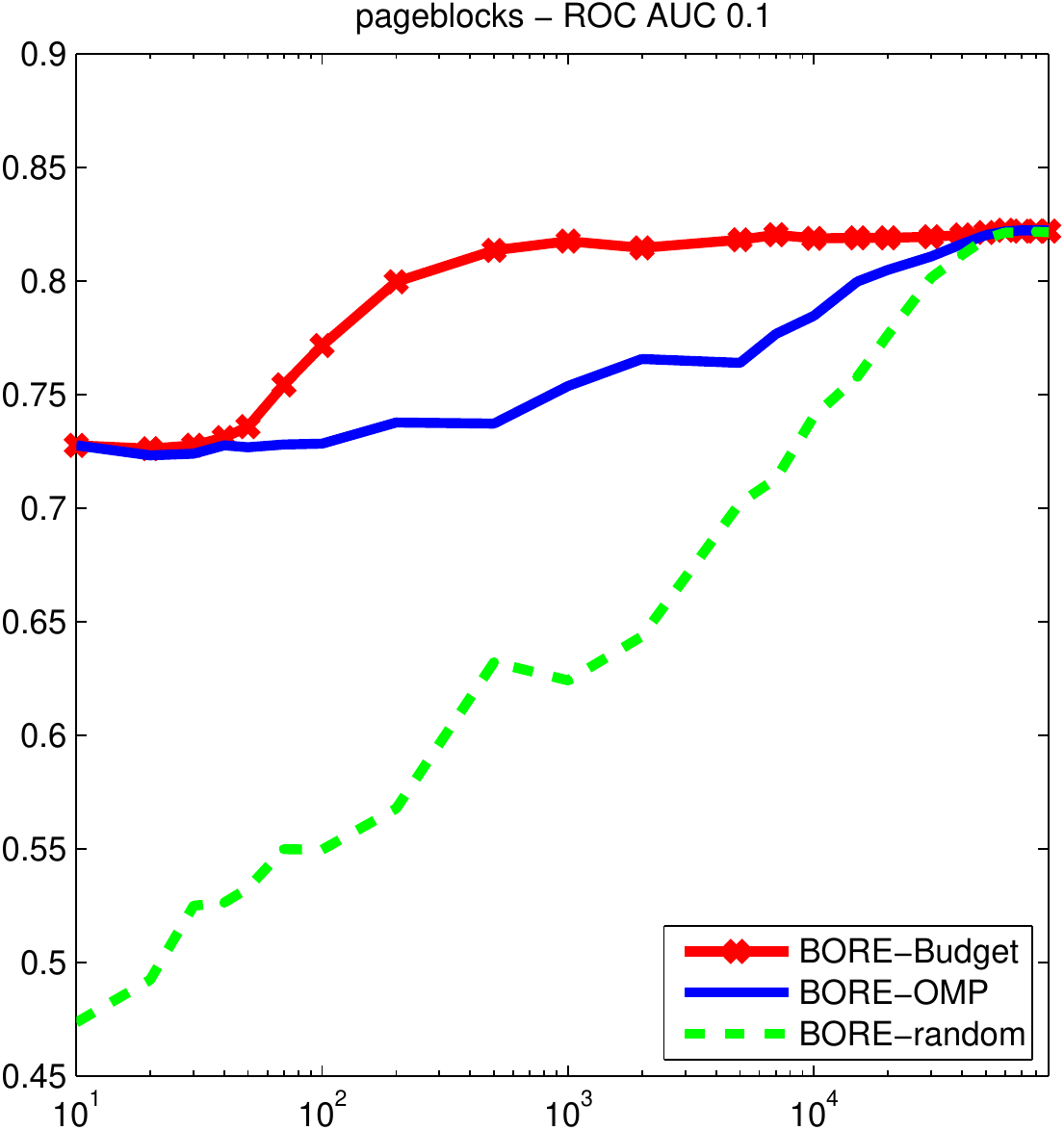}
\subcaption{Pageblocks.} \label{sub:page}
\end{subfigure} 
\begin{subfigure}[b]{0.49\textwidth}	
\includegraphics[width=0.49\textwidth]{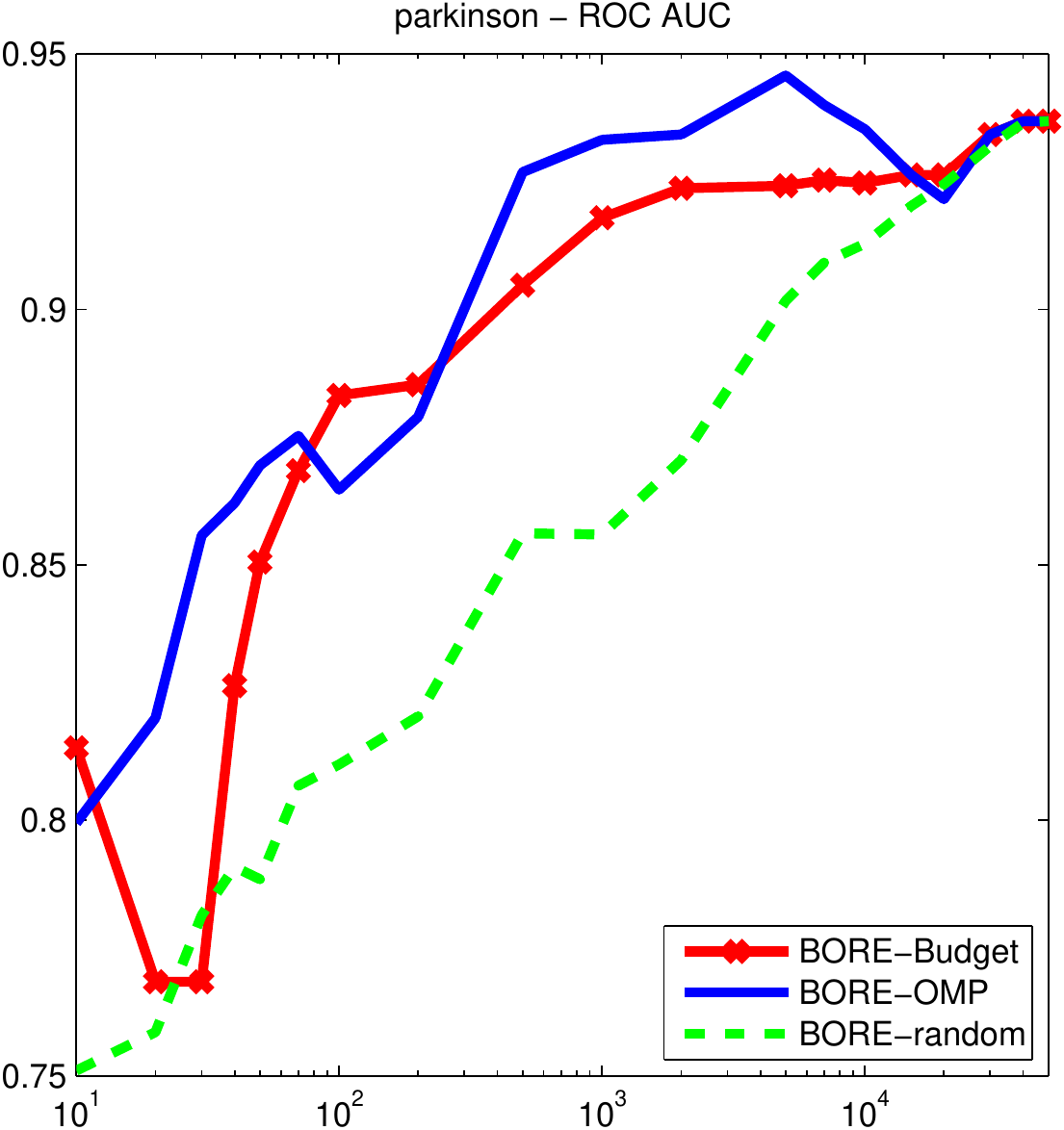}
\includegraphics[width=0.49\textwidth]{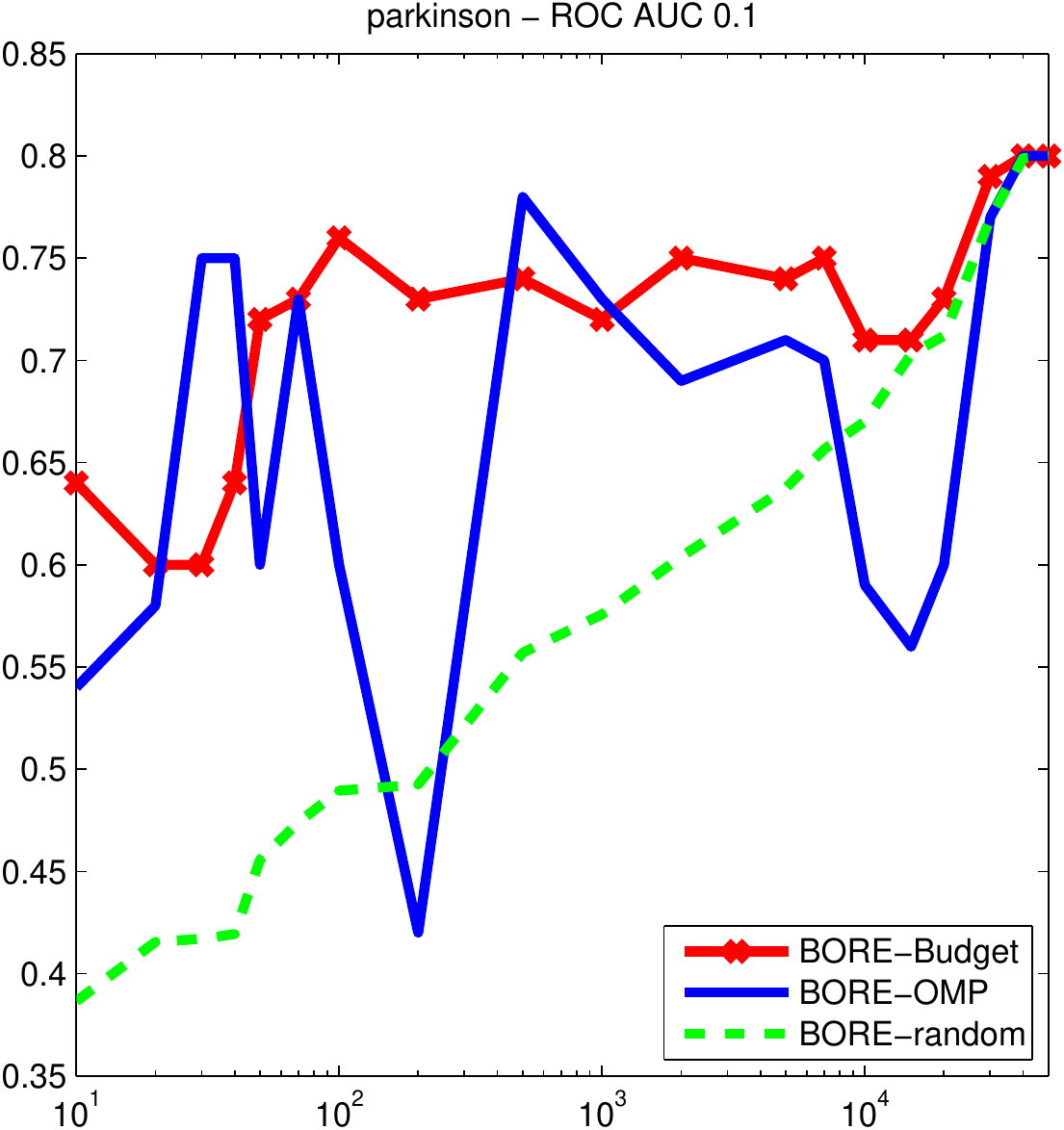}
\subcaption{Parkinson.} \label{sub:park}
\end{subfigure}
\begin{subfigure}[b]{0.49\textwidth}	
\includegraphics[width=0.49\textwidth]{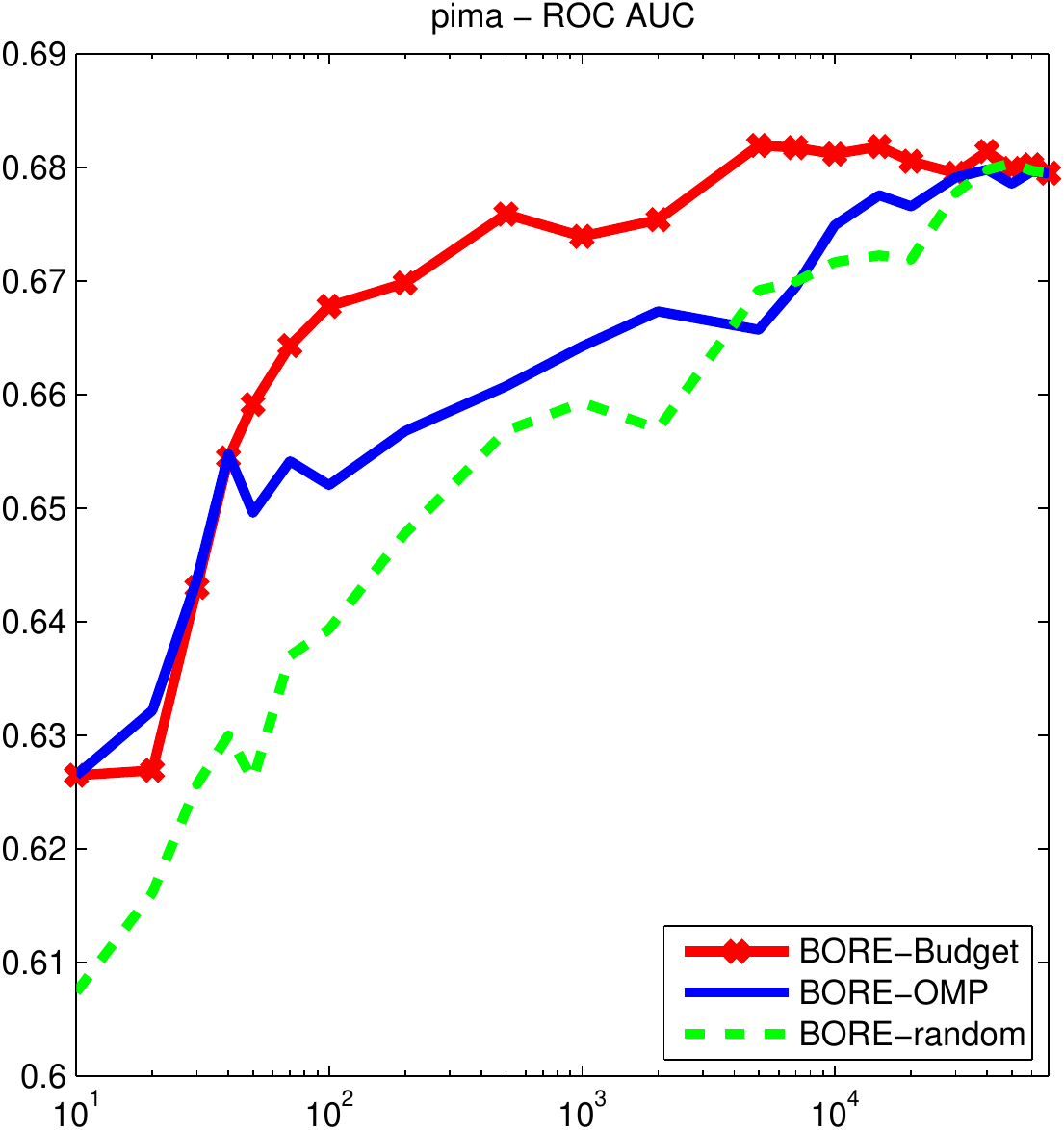}
\includegraphics[width=0.49\textwidth]{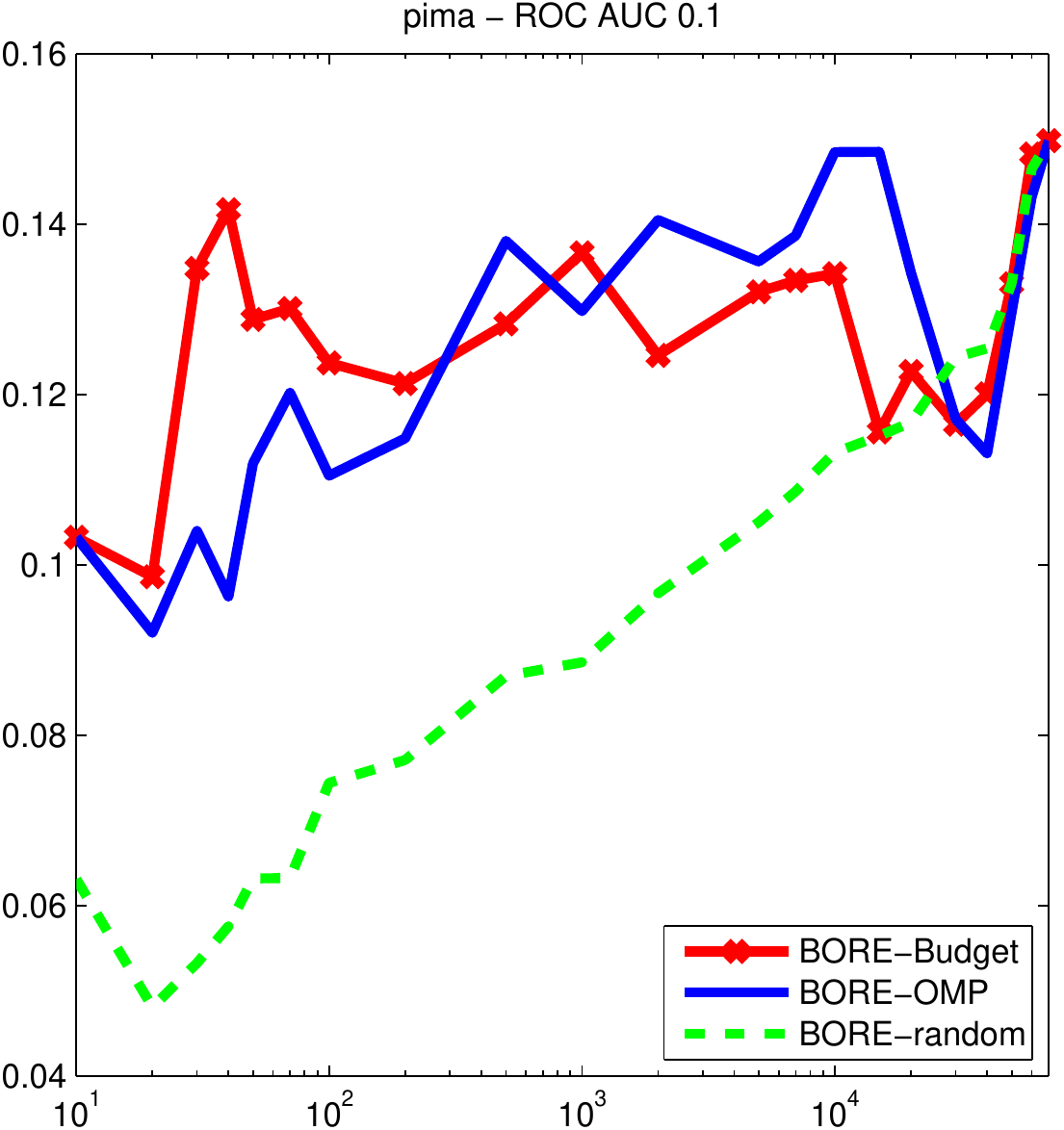}
\subcaption{Pima.} \label{sub:pima}
\end{subfigure}
\begin{subfigure}[b]{0.49\textwidth}	
\includegraphics[width=0.49\textwidth]{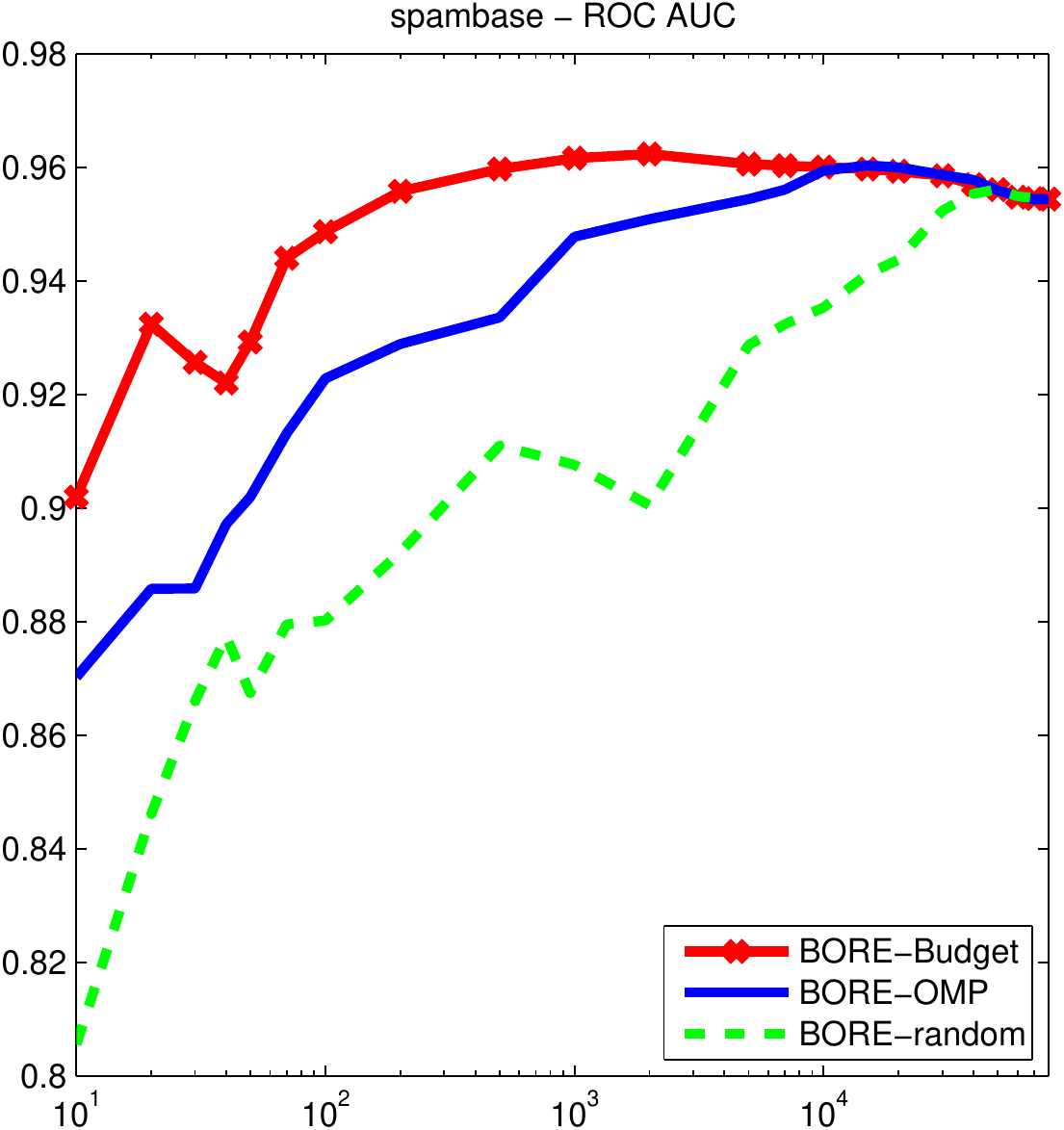}
\includegraphics[width=0.49\textwidth]{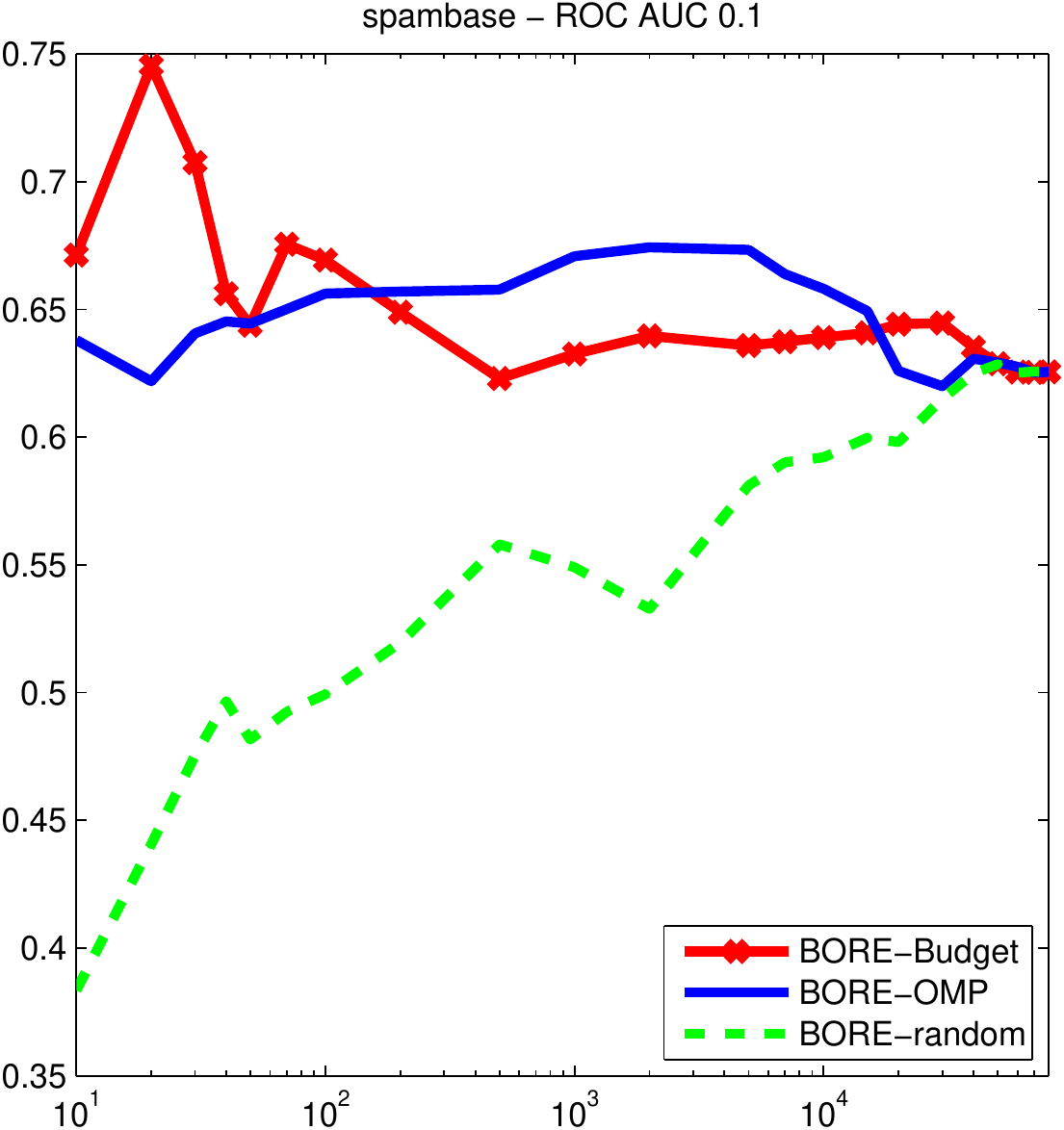}
\subcaption{Spambase.} \label{sub:spam}
\end{subfigure}
\begin{subfigure}[b]{0.49\textwidth}	
\includegraphics[width=0.49\textwidth]{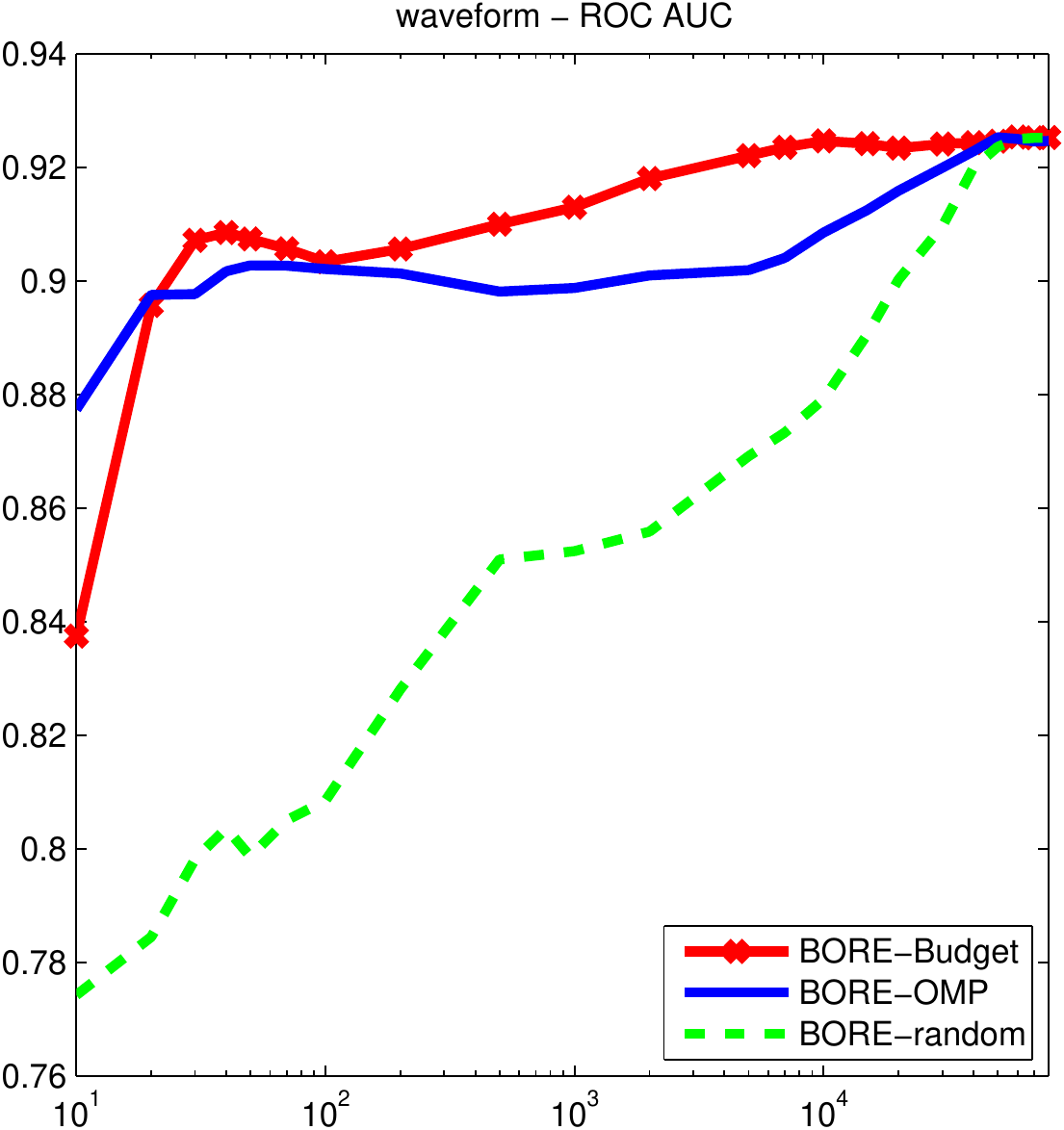}
\includegraphics[width=0.49\textwidth]{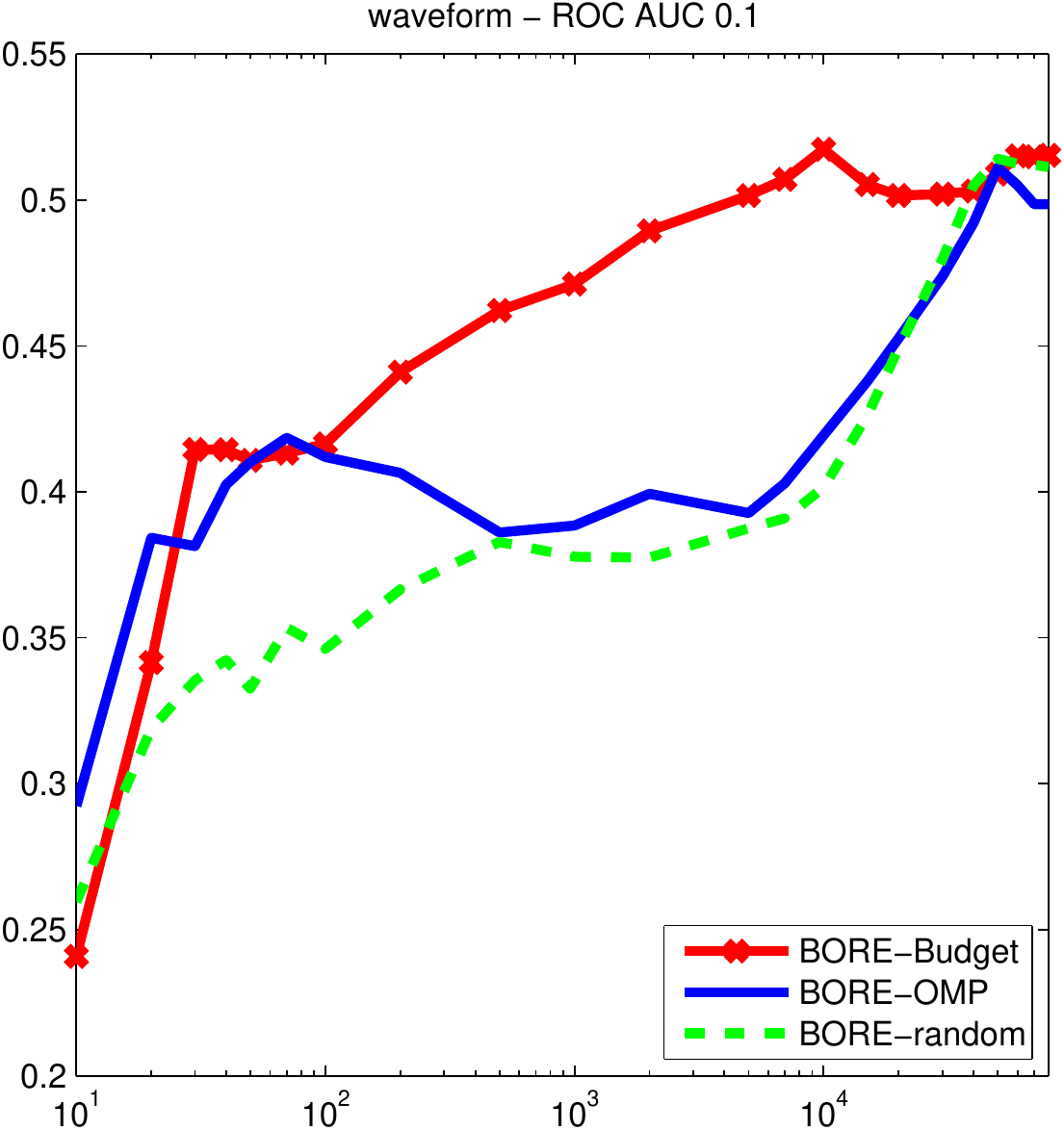}
\subcaption{Waveform.} \label{sub:wave}
\end{subfigure}
\begin{subfigure}[b]{0.49\textwidth}	
\includegraphics[width=0.49\textwidth]{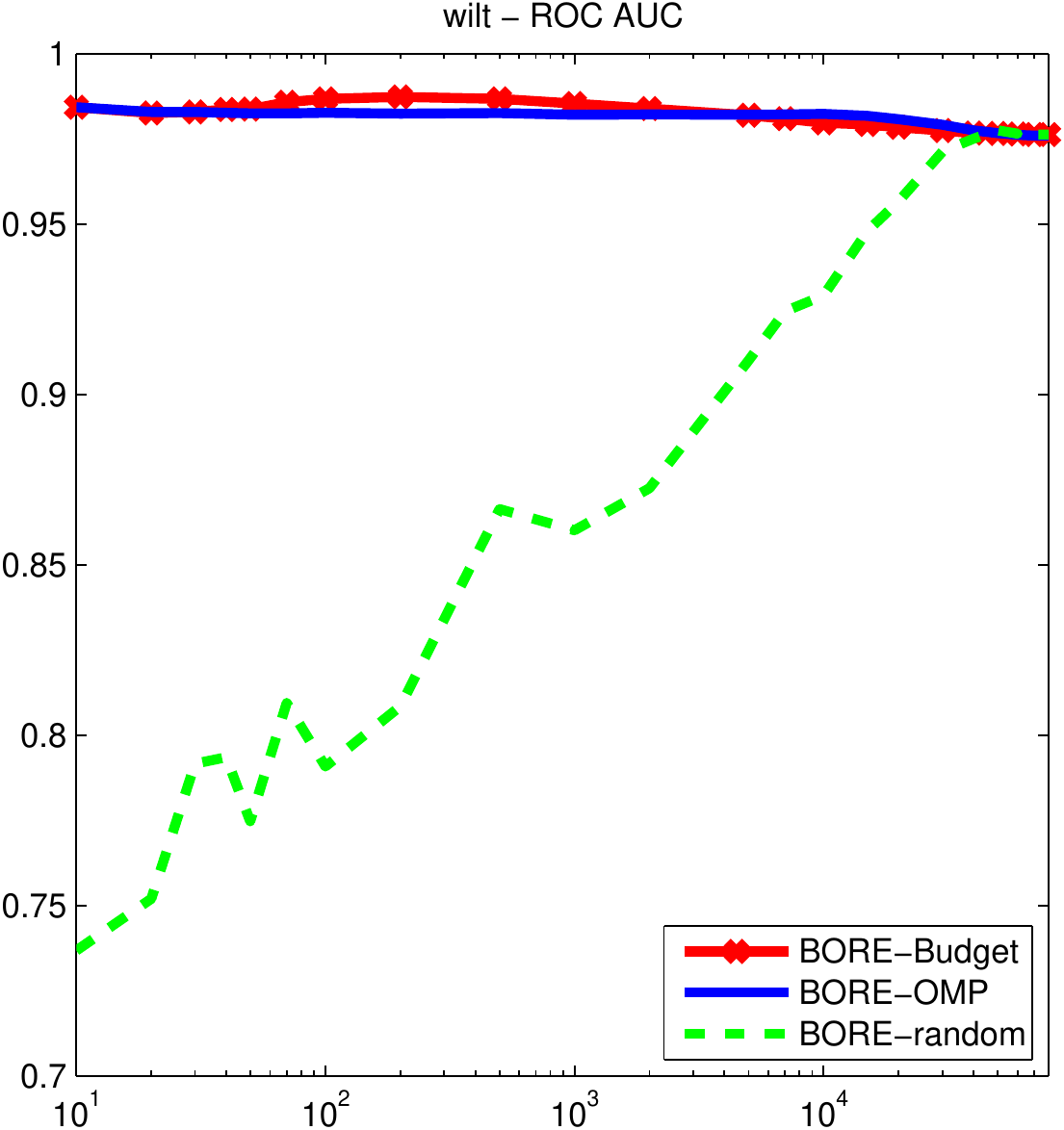}
\includegraphics[width=0.49\textwidth]{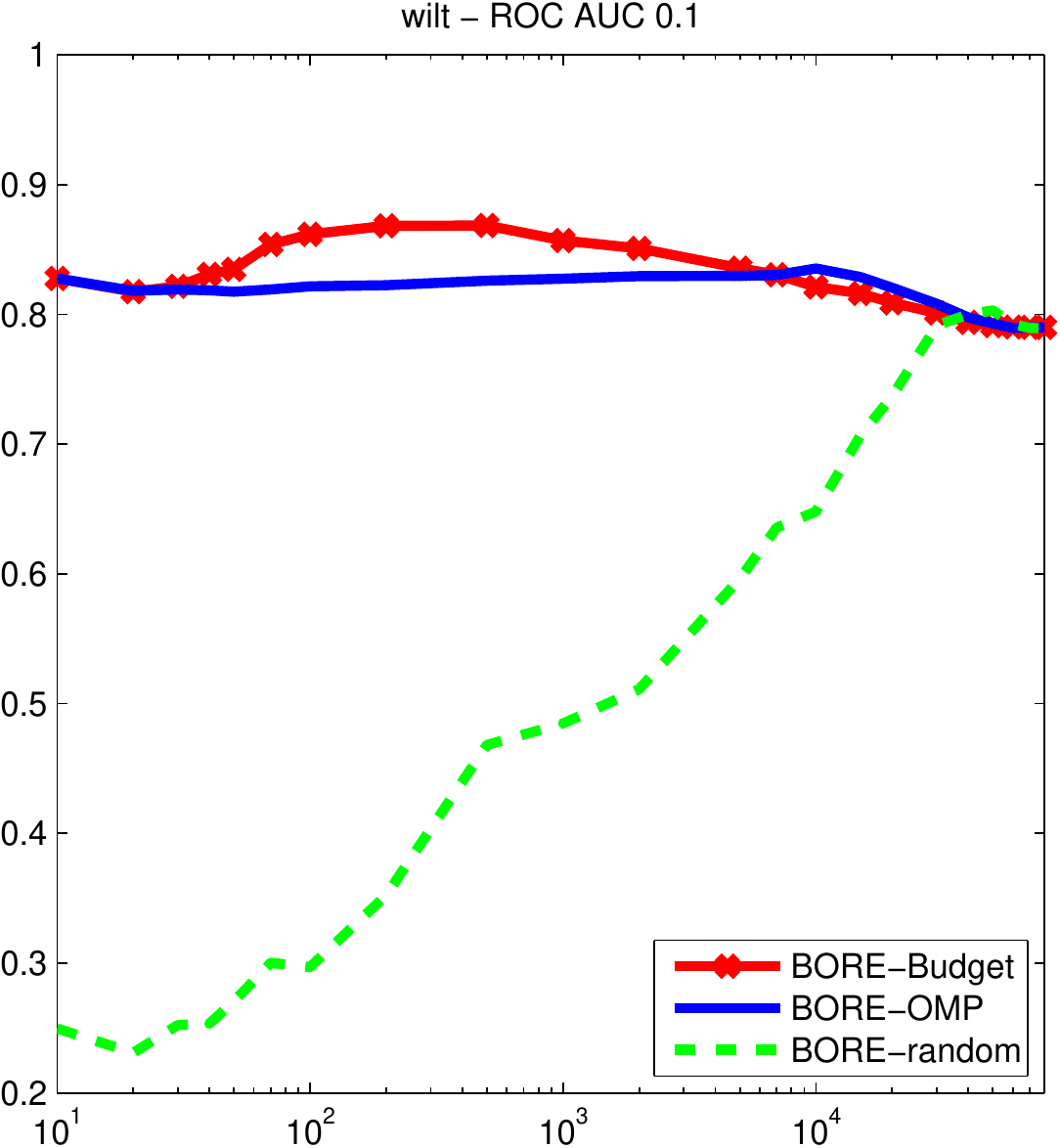}
\subcaption{Wilt.} \label{sub:wilt}
\end{subfigure}
\caption{Evaluation of outlier detection on budget. The spikes in performance are due to using a single \emph{stable} set of features, $\mathcal{S}_C$ across all bags which could differ greatly from the individual active sets in each bag $\mathcal{A}^{(b)}$ due to randomness introduced by subsampling. The smoother performance of BORE-Budget can be explained by the cost-aware feature selection procedure ensuring that since features are selected based on both cost and utility, the stable set is more similar to the active sets in each bag. Smoother performance might be obtained by downweighting the contribution of bags whose active sets differ too greatly from $\mathcal{S}_C$ instead of simple averaging. }
\label{fig:budget_eval2}
\vspace{3cm}
\end{center}
\end{figure*}

\subsection{Budgeted Outlier Detection}
\subsubsection{Algorithms and Experimental Setup} 
We now evaluate the proposed feature selection technique for BORE in the setting where a prediction-time budget is imposed. 
We report ROC AUC and AUC under the beginning of the ROC curve (AUC 0.1). Due to space limitations we omit the results for precision@$n_o$ but note that this measure exhibits similar trends to AUC $0.1$.

We compare our proposed cost-aware feature selection scheme BORE-{Budget} against BORE-{OMP} which uses the unmodified Algorithm \ref{alg:OMP} for feature selection. Note that none of the competitors from the previous experiments are designed to take into account feature costs. Therefore, as a baseline we show the average detection performance of BORE using a random subset of features selected such that the budget constraint is satisfied (BORE-{random}).

To every OSF, we assign a feature cost. We then evaluate the detection performance of the three algorithms for a series of budgets. We sample feature costs uniformly with replacement from a pool of values and randomly assign them to the features.
The pool is constructed such that the costs correspond to different computational complexities: $\{n$, $2n$, $5n$, $n^2$, $2n^2$, $3n^2$, $n^3$, $2n^3\}$. Concretely, we instantiate the pool as $\{$10, 20, 50, 100, 200, 300, 1000, 2000$\}$ for all data sets.
We report on average performance over 20 different random assignments of costs to features.
The original features always have a uniform cost of $1$ in our experiments. 

Having assigned costs, we apply OMP and budgeted OMP to get a feature ranking. We then evaluate outlier detection performance for budgets ranging from 10 (the minimal cost of a transformed feature) to the maximum which is the sum of all feature costs. For every budget, each method selects a set of features such that the budget constraints are satisfied. For the baseline BORE-random, we randomly select a subset of features which satisfy the budget constraint and report average performance over 20 random subsamples.

The number of bags and outlier subsampling ratio for BORE are the same as for the previous experiments ($50$ and $70\%$, respectively).


\subsubsection{Results}
Results are shown in Fig.\,\ref{fig:budget_eval1}. 
In 9 of the 12 datasets, for any given budget BORE-Budget typically achieves a larger AUC and AUC $0.1$ than the competing methods. This is expected since BORE-Budget selects features taking their cost into account. 

In the setting with a single bag, the AUC is expected to increase monotonically \cite{grubb2014anytime} -- i.e.\ as the budget increases, the performance of the method should always improve. The lack of sctrict monotonicity exhibited by BORE-Budget and BORE-OMP is explained due to the effects of bagging and stability selection which are necessary to provide good performance in the highly imbalanced setting. However BORE-Budget exhibits \emph{smoother} behaviour than the competitors as the budget changes. 

The poor performance of random selection underlines the importance of a principled feature selection procedure. Interestingly, the relatively monotonic behaviour of the random baseline as more features are added to the model underscores that BORE benefits from using a larger number of OSFs as feature transformations.  This further emphasises that in the absence of strict computational constraints a large and diverse set of OSFs should be chosen to construct the outlier representation.

%

\section{Conclusion} \label{sec-dis}

We have introduced BORE, an approach to outlier detection which combines unsupervised and supervised techniques in order to build a rich representation of the outliers in the data. One of the main benefits of BORE is its simplicity, which takes advantage of decades of existing research in designing outlier scoring functions to result in a powerful algorithm which is insensitive to tuning parameters. 
BORE is based on effective supervised learning methods that are well studied, and leverages the recent wisdom that learning a good \emph{representation} is of utmost importance to training a simple yet highly predictive model. In this manner, we propose an entirely new way of integrating unsupervised information into supervised outlier detection. We have shown that BORE outperforms existing unsupervised and supervised methods on a wide range of real world datasets.

Another key benefit of the BORE framework is its generality and extendibility. 
For example, newly developed OSFs can easily be incorporated as part of the feature representation. Specific domain knowledge can also be encoded implicitly by the choice of OSFs. Furthermore, existing supervised outlier detection techniques could be complemented using the BORE framework. This removes some of the guesswork inherent in designing non-linear feature transformations (i.e. kernels) for 
\afterpage{\clearpage}
specific tasks. We have tested this hypothesis empirically with SSAD+OR and we have observed improved performance over standard SSAD for half of the data.

In the context of recent research on outlier ensembles \cite{Agg-posPaper}, BORE can be viewed as the first supervised ensemble technique. It learns weights for OSFs and depending on the specific classifier used, its final output can be easily interpreted as a probability. In contrast, existing unsupervised outlier ensembles struggle with proper normalization of the OSFs outputs and are thus difficult to interpret 
\cite{KriKroSchZim11,SchWojZimKri12}. BORE avoids this problem by learning appropriate thresholds between inliers and outliers from the training data.

Finally, we concentrate on the problem of reducing the computational cost at test-time. 
We envisage a scenario where resources at training time are plentiful. For example, the OSFs can be trained in parallel (which is only required once for the full dataset) as can the supervised models on each subsampled bag of data. For context, most successful approaches to representation learning require large amounts of resources at training \emph{and} test time (i.e. deep networks).
BORE is the only method capable of handling budget constraints at test time. We have shown that it successfully selects a subset of features that provide good overall outlier detection performance.



\bibliographystyle{abbrv}
\bibliography{bore_ICDM}

\begin{thebibliography}{10}

\bibitem{Agg-posPaper}
C.~C. Aggarwal.
\newblock Outlier ensembles: position paper.
\newblock {\em SIGKDD Explorations}, 14(2), 2012.

\bibitem{Agg13}
C.~C. Aggarwal.
\newblock {\em Outlier Analysis}.
\newblock Springer, 2013.

\bibitem{AngPiz05}
F.~Angiulli and C.~Pizzuti.
\newblock Outlier mining in large high-dimensional data sets.
\newblock 17(2):203--215, 2005.

\bibitem{uci}
K.~Bache and M.~Lichman.
\newblock {UCI} machine learning repository, 2013.

\bibitem{BarLew94}
V.~Barnett and T.~Lewis.
\newblock {\em Outliers in Statistical Data}.
\newblock 3rd edition, 1994.

\bibitem{bengio2013representation}
Y.~Bengio, A.~Courville, and P.~Vincent.
\newblock Representation learning: A review and new perspectives.
\newblock {\em Pattern Analysis and Machine Intelligence, IEEE Transactions
  on}, 35(8):1798--1828, 2013.

\bibitem{Breiman-bag}
L.~Breiman.
\newblock Bagging predictors.
\newblock {\em Machine Learning}, 24(2), 1996.

\bibitem{breiman1996stacked}
L.~Breiman.
\newblock Stacked regressions.
\newblock {\em Machine learning}, 24(1):49--64, 1996.

\bibitem{BreKriNgSan00}
M.~M. Breunig, H.-P. Kriegel, R.~Ng, and J.~Sander.
\newblock {LOF}: Identifying density-based local outliers.
\newblock In {\em SIGMOD'00}, 2000.

\bibitem{buhlmann2014magging}
P.~B{\"u}hlmann and N.~Meinshausen.
\newblock Magging: maximin aggregation for inhomogeneous large-scale data.
\newblock {\em arXiv preprint arXiv:1409.2638}, 2014.

\bibitem{ChaBanKum09}
V.~Chandola, A.~Banerjee, and V.~Kumar.
\newblock Anomaly detection: A survey.
\newblock {\em CSUR}, 41(3), 2009.

\bibitem{jair2013-semisup}
N.~Goernitz, M.~Kloft, K.~Rieck, and U.~Brefeld.
\newblock Toward supervised anomaly detection.
\newblock {\em J. Artif. Intell. Res. (JAIR)}, 46, 2013.

\bibitem{grubb2014anytime}
A.~Grubb.
\newblock {\em Anytime Prediction: Efficient Ensemble Methods for Any
  Computational Budget}.
\newblock PhD thesis, University of Maryland, 2014.

\bibitem{hastie2009elements}
T.~Hastie, R.~Tibshirani, J.~Friedman, T.~Hastie, J.~Friedman, and
  R.~Tibshirani.
\newblock {\em The elements of statistical learning}, volume~2.
\newblock Springer, 2009.

\bibitem{HauKaeFra04}
V.~Hautam\"{a}ki, I.~K\"{a}rkk\"{a}inen, and P.~Fr\"{a}nti.
\newblock Outlier detection using k-nearest neighbor graph.
\newblock In {\em Proc. od the 14th Int. Conf. on Pattern Recognition},
  ICPR'04, pages 430--433, 2004.

\bibitem{JinTunHanWan06}
W.~Jin, A.~K.~H. Tung, J.~Han, and W.~Wang.
\newblock Ranking outliers using symmetric neighborhood relationship.
\newblock In {\em PAKDD'06}, pages 577--593, 2006.

\bibitem{Jolliffe1986}
I.~Jolliffe.
\newblock {\em Principal Component Analysis}.
\newblock 1986.

\bibitem{KriKroSchZim09a}
H.-P. Kriegel, P.~Kr\"{o}ger, E.~Schubert, and A.~Zimek.
\newblock {LoOP:} local outlier probabilities.
\newblock In {\em CIKM'09}, pages 1649--1652, 2009.

\bibitem{KriKroSchZim11}
H.-P. Kriegel, P.~Kr\"oger, E.~Schubert, and A.~Zimek.
\newblock Interpreting and unifying outlier scores.
\newblock In {\em SDM'11}, 2011.

\bibitem{KriSchZim08}
H.-P. Kriegel, M.~Schubert, and A.~Zimek.
\newblock Angle-based outlier detection in high-dimensional data.
\newblock In {\em KDD'08}, pages 444--452, 2008.

\bibitem{LatLazPok07}
L.~J. Latecki, A.~Lazarevic, and D.~Pokrajac.
\newblock Outlier detection with kernel density functions.
\newblock In {\em MLDM'07}, pages 61--75, 2007.

\bibitem{LazKum05}
A.~Lazarevic and V.~Kumar.
\newblock Feature bagging for outlier detection.
\newblock In {\em KDD'05}, 2005.

\bibitem{lozano2011group}
A.~C. Lozano, G.~Swirszcz, and N.~Abe.
\newblock Group orthogonal matching pursuit for logistic regression.
\newblock In {\em Int. Conf. on Artif. Intelligence and Statistics}, 2011.

\bibitem{mcwilliams2013correlated}
B.~McWilliams, D.~Balduzzi, and J.~Buhmann.
\newblock Correlated random features for fast semi-supervised learning.
\newblock In {\em Advances in Neural Information Processing Systems}, pages
  440--448, 2013.

\bibitem{Meinshausen2010}
N.~Meinshausen and P.~B\"{u}hlmann.
\newblock Stability selection.
\newblock {\em Journal of the Royal Stat. Society. Series B}, 72(4), 2010.

\bibitem{learnOutEns}
B.~Micenkov\'{a}, B.~McWilliams, and I.~Assent.
\newblock Learning outlier ensembles: The best of both worlds - supervised and
  unsupervised.
\newblock In {\em Proc. of the ACM SIGKDD Workshop on Outlier Detection and
  Description}, ODD'14, 2014.

\bibitem{MSS11}
E.~M\"{u}ller, M.~Schiffer, and T.~Seidl.
\newblock Statistical selection of relevant subspace projections for outlier
  ranking.
\newblock In {\em Proc. of the 27th Int. Conf. on Data Engineering}, ICDE'11,
  pages 434--445, 2011.

\bibitem{nguyen_ensemble}
H.~V. Nguyen, H.~H. Ang, and V.~Gopalkrishnan.
\newblock Mining outliers with ensemble of heterogeneous detectors on random
  subspaces.
\newblock In {\em Proc. of the 15th Int. Conf. on Database Syst. for Adv.
  Appl.}, DASFAA'10, 2010.

\bibitem{RamRasShi00}
S.~Ramaswamy, R.~Rastogi, and K.~Shim.
\newblock Efficient algorithms for mining outliers from large data sets.
\newblock In {\em Proc. of the 19th Int. Conf. on Management of Data},
  SIGMOD'00, pages 427--438, 2000.

\bibitem{SchWojZimKri12}
E.~Schubert, R.~Wojdanowski, A.~Zimek, and H.-P. Kriegel.
\newblock On evaluation of outlier rankings and outlier scores.
\newblock In {\em SDM'12}, 2012.

\bibitem{SchZimKri14a}
E.~Schubert, A.~Zimek, and H.-P. Kriegel.
\newblock Generalized outlier detection with flexible kernel density estimates.
\newblock In {\em SDM'14}, pages 542--550, 2014.

\bibitem{SchZimKri14}
E.~Schubert, A.~Zimek, and H.-P. Kriegel.
\newblock Local outlier detection reconsidered: a generalized view on locality
  with applications to spatial, video, and network outlier detection.
\newblock 28(1):190--237, 2014.

\bibitem{TanCheFuChe02}
J.~Tang, Z.~Chen, A.~W.-C. Fu, and D.~W. Cheung.
\newblock Enhancing effectiveness of outlier detections for low density
  patterns.
\newblock In {\em PAKDD'02}, pages 535--548, 2002.

\bibitem{svdd}
D.~M.~J. Tax and R.~P.~W. Duin.
\newblock Support vector data description.
\newblock {\em Mach. Learn.}, 54(1), 2004.

\bibitem{van2008visualizing}
L.~Van~der Maaten and G.~Hinton.
\newblock Visualizing data using t-sne.
\newblock {\em Journal of Machine Learning Research}, 9(2579-2605):85, 2008.

\bibitem{VuGA11}
N.~H. Vu, V.~Gopalkrishnan, and I.~Assent.
\newblock An unbiased distance-based outlier detection approach for
  high-dimensional data.
\newblock In {\em DASFAA (1)}, volume 6587, pages 138--152, 2011.

\bibitem{combat-imbal}
M.~Wasikowski and X.~wen Chen.
\newblock Combating the small sample class imbalance problem using feature
  selection.
\newblock {\em IEEE Trans. on Knowledge and Data Engineering}, 22(10), 2010.

\bibitem{ZhaHutJin09}
K.~Zhang, M.~Hutter, and H.~Jin.
\newblock A new local distance-based outlier detection approach for scattered
  real-world data.
\newblock In {\em PAKDD'09}, pages 813--822, 2009.

\bibitem{Zimek-posPaper}
A.~Zimek, R.~J. G.~B. Campello, and J.~Sander.
\newblock Ensembles for unsupervised outlier detection: challenges and research
  questions a position paper.
\newblock {\em SIGKDD Explorations}, 15(1), 2013.

\end{thebibliography}

\end{document}